\documentclass[11pt]{article}

\usepackage[margin=1in]{geometry}
\usepackage[T1]{fontenc}
\usepackage[utf8]{inputenc}
\usepackage{lmodern}
\usepackage{microtype}
\usepackage{amsmath,amssymb}
\usepackage{graphicx}
\usepackage{booktabs}
\usepackage{makecell}
\usepackage{caption}
\usepackage[round]{natbib}
\usepackage{xcolor}
\usepackage{hyperref}
\hypersetup{colorlinks=true, linkcolor=blue!50!black, citecolor=blue!50!black, urlcolor=blue!50!black}

\graphicspath{{figures/}}
\newcommand{\muV}{\ensuremath{\mu}V}

\title{Decoding silent reading from non-invasive EEG}

\author{Ingo Marquardt\thanks{Corresponding author: \texttt{ingo@nubrain.com}}, Anthilia Alchanat, Priyanka Jain\\[0.5em] nubrain}

\date{18 August 2026}

\begin{document}

\maketitle

\begin{abstract}
Non-invasive decoding of inner speech faces a fundamental data problem: a corpus pairing brain activity with a person's spontaneous inner monologue cannot be collected, and the available proxy paradigms (cued \emph{repetitive} and retrospectively reported \emph{generative} inner speech) are slow to acquire, poorly time-locked, and subject compliance is unverifiable. We therefore treat silent reading as a scalable proxy task and ask how much lexical and semantic information a contrastive decoder can extract from it. We report an open-vocabulary analysis of approximately 240,000 word presentations recorded from a single densely-sampled participant across 393 runs (ca. 49 h) of 19-channel dry-electrode EEG. Words from continuous narrative text were presented in rapid serial visual presentation, with typography randomised on every trial to partially decorrelate word identity from low-level visual form. A convolutional EEG encoder, optionally followed by a causal transformer, was trained with a CLIP-style contrastive objective to align short EEG windows with hidden-state embeddings of the presented word taken from a large language model. Decoding, evaluated as word-grouped top-10 retrieval against permutation baselines, was reliably above chance, extended to mid-frequency and rare words, and scaled log-linearly with training-data volume with no sign of saturation. Removing occipital and posterior-temporal electrodes reduced the word-level gain by roughly one third but left context tracking unchanged. Control analyses separate word-level decoding from narrative context tracking and from a non-neural positional prior introduced by the transformer's positional embedding. These results establish that open-vocabulary word-level information is recoverable from EEG during silent reading, and that decoding is data-limited rather than saturated.
\end{abstract}

\section{Introduction}
\label{sec:introduction}

\subsection{Motivation}
\label{sec:motivation}

Loss of speech following stroke, traumatic brain injury, amyotrophic lateral sclerosis or other neurodegenerative disease removes the most basic channel of human communication. Brain-computer interfaces (BCIs) that decode intended or imagined speech directly from neural activity have therefore been a long-standing clinical goal, and the last five years have delivered striking progress, but mostly using invasive techniques. Intracortical and electrocorticographic systems now decode attempted speech at conversational rates and near-perfect accuracy over vocabularies of tens of thousands of words \citep{card2024accurate, metzger2023high, moses2021neuroprosthesis}, and recent work has shown that \emph{inner} speech --- imagined speech with no attempted articulation --- is robustly represented in motor cortex as a scaled-down version of attempted speech occupying a shared neural subspace, permitting real-time decoding of imagined sentences from a 125{,}000-word vocabulary \citep{kunz2025inner}. Single-neuron recordings in supramarginal gyrus tell a similar story, with internal speech decodable online and with representations shared across reading, listening, and vocalised speech \citep{wandelt2024representation}.

These results settle the scientific question of whether inner speech can be decoded from measured brain activity. What they do not settle is whether that structure is accessible without neurosurgery. Invasive devices carry surgical risk, have limited chronic longevity, and will not scale to the population of people who would benefit from restored or augmented communication. The open question is whether non-invasive recordings, and specifically EEG, can be pushed far enough to be useful.

\subsection{The state of non-invasive language decoding}
\label{sec:state-of-field}

The non-invasive language decoding literature divides into two distinct approaches with very different outcomes.

The first approach decodes language that the participant \emph{perceives}. \citet{defossez2023decoding} trained a convolutional brain module with a contrastive (CLIP) objective to align M/EEG with self-supervised speech representations from wav2vec 2.0, and identified the correct 3-second speech segment out of more than a thousand candidates with up to 41\% top-1 accuracy from MEG. However, their EEG results were far weaker (top-10 accuracy of 17.7\% and 25.7\% on two datasets). \citet{sato2024scaling} recorded a staggering 175\,h of EEG from a single participant reading aloud, and reached 48.5\% top-1 and 76.0\% top-10 classification accuracy over 512 candidate segments, and, more importantly than the headline number, demonstrated an unsaturated log-linear scaling relationship between data volume and decoding accuracy. When they subsampled their data to the ${\sim}3$\,h typical of the field, accuracy collapsed to near the levels previously reported. Their conclusion, which motivates the present study, is that the binding constraint on EEG language decoding may be dataset size rather than a hard ceiling on signal quality.

The second approach to non-invasive language decoding attempts to decode \emph{inner} speech directly, and the results are markedly worse. \citet{csaky2025towards} collected an unusually large single-participant EEG and MEG dataset across three paradigms --- silent reading, repetitive inner speech, and generative inner speech --- using a five-word vocabulary chosen for clinical relevance. Silent reading decoded at 30--40\% (chance 20\%) across EEG, MEG and optically-pumped magnetometers, with permutation feature importance localising the effect to visual cortex at roughly 150\,ms. Both inner-speech paradigms were essentially at chance, across a wide range of decoding methods, and the failure prevented any test of transfer from reading to inner speech. Earlier EEG studies of imagined speech report accuracies that are statistically above chance but far from usable, typically on small vocabularies and few electrodes \citep{cooney2019classification, simistira2022rethinking}, with occasional outliers whose paradigms invite confounds \citep{abdulghani2024enhancing}. Direct MEG investigations of imagined phrases have reported high accuracies \citep{dash2020decoding}, but with paradigms in which imagination immediately follows perception of the same phrase within a trial, leaving residual perceptual and motor-preparatory activity as plausible contributors.

\subsection{The labelled-data problem for inner monologue}
\label{sec:labelled-data-problem}

The gap between these two approaches to non-invasive language decoding is instructive. Where large volumes of well-time-locked data are available (as in perception and overt production), non-invasive decoding works, imperfectly but measurably. Where they are not (spontaneous inner speech), decoding does not work.

This discrepancy reflects a fundamental obstacle. The ideal training corpus for an inner-monologue decoder would pair short EEG segments with a transcription of what the person was thinking during that segment. Such a corpus cannot exist. To know a person's inner monologue at a precise point in time, one needs a device that reads minds; to build that device one needs the corpus. Every existing paradigm is a way of partially escaping this circle, and each pays a specific price.

\emph{Generative inner speech}, in which participants freely imagine an item and report it afterwards \citep{jones2021note, csaky2025towards}, preserves the endogenous character of the thought but destroys its timing. Even a truthful report of having thought ``I like bananas'' does not tell us whether the thought occurred 400\,ms or 1{,}500\,ms before the report; a fatal ambiguity for a signal whose informative structure lies on a scale of tens of milliseconds. It also demands sustained, unverifiable compliance: there is no objective check on whether a participant was on task, mind-wandering, or reporting selectively. Data quality and data quantity both suffer.

\emph{Repetitive inner speech}, in which participants are instructed which item to imagine and cued when to imagine it, restores timing but reintroduces the cue. The recorded response contains the neural processing of the cue as well as the imagined item. Careful designs mitigate this. Presenting the item briefly, waiting a jittered interval, then delivering a content-neutral go signal reduces sensory contamination, but does not eliminate it. (Perception-based proxy tasks, like the one in the present study, are obviously also affected by sensory signals, as discussed in Section~\ref{sec:visual-confound}.)

But in our view, even more important than a lack of precise temporal information (in generative inner speech) and sensory contamination (in repetitive inner speech) is the fact that these paradigms impose a slow trial structure. Slow, repetitive trials are boring, bored participants disengage, and disengagement is undetectable in the data. The paradigm is therefore doubly limited: it does not scale, and its labels degrade in exactly the way that cannot be audited. Neither approach can plausibly produce the tens or hundreds of hours per participant that the scaling results of \citet{sato2024scaling} suggest are necessary.

\subsection{Our approach: reading (and, later, listening) as a scalable proxy}
\label{sec:our-approach}

Any language decoding paradigm must trade off three objectives that pull against each other:

\begin{enumerate}
  \item \textbf{Validity}: The task must induce language processing and activate semantic representations. For example, a task that can be performed by paying attention to visual properties of word stimuli does not guarantee semantic processing.
  \item \textbf{Scalability}: The paradigm must cover a large amount of linguistic content per hour and remain viable over several hours.
  \item \textbf{Compliance}: The task must be engaging enough to sustain attention, easy enough to follow without excessive mental effort, and instrumented so that lapses can be detected.
\end{enumerate}

We concluded that the best available compromise for obtaining a training dataset for language decoding is a multimodal paradigm combining silent reading and passive listening. When a word is read, early visual cortex first encodes letter shapes and positions. Once the word form is recognised, higher-level areas are engaged in encoding its meaning. When the same word is heard, auditory cortex first encodes the phonetic structure, but the same semantic representation is ultimately evoked. Reading and listening thus share a representational target while differing maximally in their input-driven, modality-specific components. A decoder trained across both, or evaluated for transfer between them, offers a remedy against the confound that limits silent-reading paradigms, namely that the decodable signal may be mostly driven by visual word-form processing \citep{csaky2025towards, ling2019visual}. Evidence for modality-general semantic codes is well established. Language-independent semantic representations in anterior temporal lobe generalise across the two languages of bilingual listeners \citep{correia2014brain}, and fMRI classifiers trained on language perception successfully decode the content of language production \citep{lin2022neural}.

The present report covers only the silent-reading half of the multimodal language decoding design. The passive-listening condition is planned for a follow-up study, and a cross-modal transfer analysis is therefore not yet possible.

Relative to previous non-invasive silent-reading work, our design differs in three respects that matter for the interpretation of the results. First, the vocabulary is open and natural. Participants read continuous narrative prose rather than a small fixed set of clinically-motivated words, so the model must operate over tens of thousands of word types rather than a handful. Second, typography is randomised trial-by-trial (font, size, colour, character spacing), directly addressing the concern raised by \citet{csaky2025towards} that low-level visual form is confounded with word identity when each word is always rendered identically. Third, the data volume is large enough to address the scaling question.

\subsection{Aims of this report}
\label{sec:aims}

This report presents results from the first stage of an ongoing research project. The present results are restricted to a single densely-sampled participant. We have separately collected one or two sessions of silent reading data across 60 participants, and plan to present cross-subject results and multimodal results (including a passive listening condition) in future reports. In the present report, we specifically ask:

\begin{enumerate}
  \item Is word-level information decodable at all from EEG during silent reading, in an open-vocabulary contrastive setting?
  \item How much of any apparent decoding is context tracking, i.e.\ following the ongoing narrative topic, rather than identifying the currently presented word? This is the central interpretive risk when target representations come from a causal language model, and we address it directly. A related risk is that a sequence model's learned positional embedding acts as a positional prior on contextual targets. (The positional embedding is the only trial-varying, non-neural input to the model.) We measure this contribution directly with masking probes.
  \item Is the signal confined to frequent (largely function) words, or does it extend to mid-frequency and rare content words?
  \item How does performance scale with training-data volume?
  \item Which analysis and optimisation choices maximise decoding accuracy, since the practical motivation for this work is BCI development rather than a purely descriptive characterisation.
\end{enumerate}

\begin{figure}[tp]
  \centering
  \includegraphics[alt={Four-panel overview of the experimental paradigm, recording montage, decoding architecture and scoring procedure},width=\textwidth]{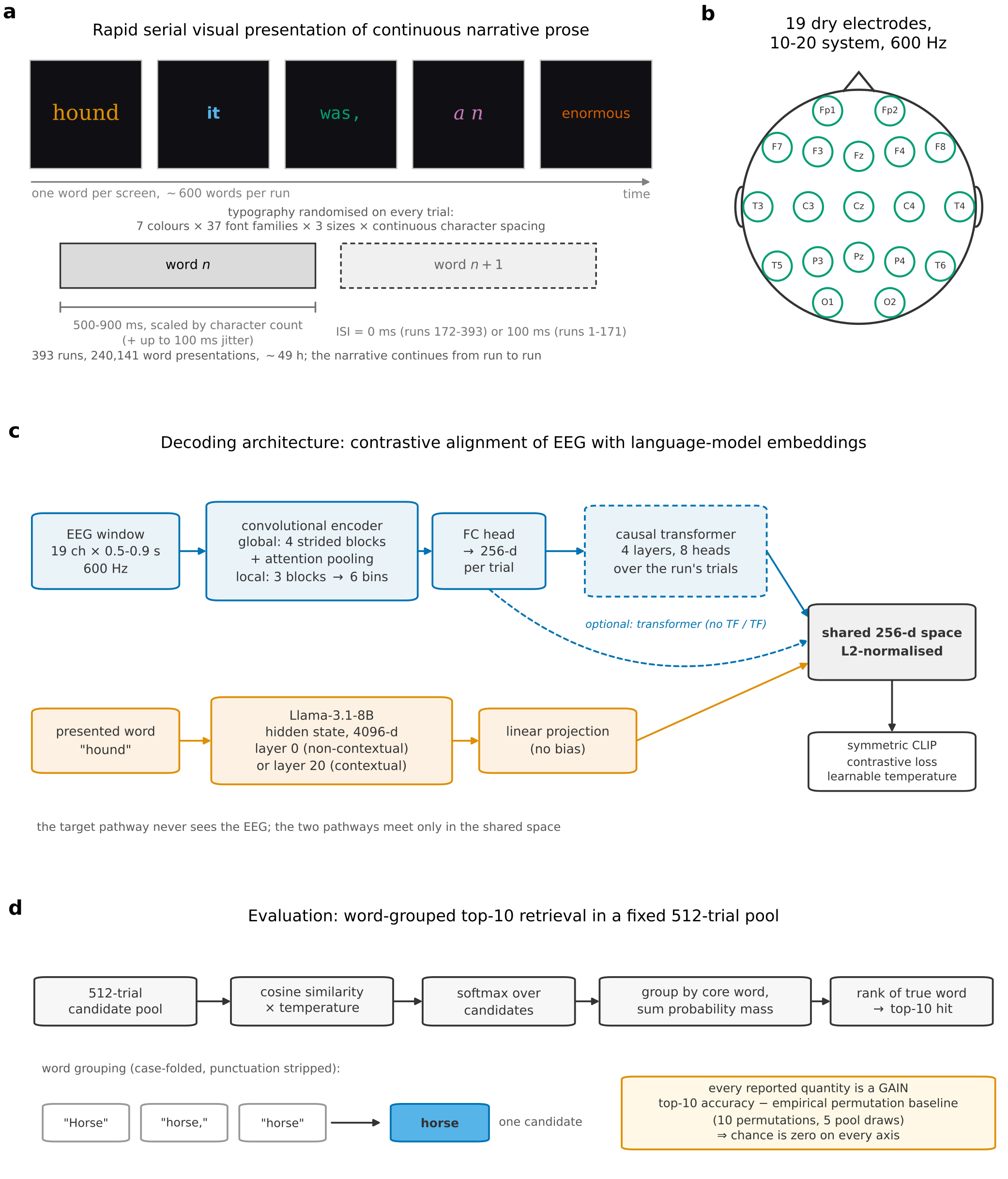}
  \caption{\textbf{Paradigm and decoding pipeline.} Full caption on the following page.}
  \label{fig:paradigm}
\end{figure}
\begin{figure}[!t]
  \caption*{\textbf{Figure~\ref{fig:paradigm} (preceding page): Paradigm and decoding pipeline.} \textbf{(a)} Words from continuous narrative prose are presented one at a time at the centre of a black screen (rapid serial visual presentation). Typography (colour, font, size, and character spacing) is drawn randomly on every trial, so that word identity is decorrelated from low-level visual form. Each word remains on screen for 500--900\,ms, scaled by character count plus up to 100\,ms of jitter. The inter-stimulus interval was 100\,ms for the deep participant's first 171 runs and 0\,ms thereafter. Runs contain ${\sim}600$ words, and the narrative continues from one run to the next. \textbf{(b)} Recording montage: 19 dry scalp electrodes in the 10-20 layout, sampled at 600\,Hz (Supplementary~\ref{sec:s0}). \textbf{(c)} Decoding architecture. A fixed-length EEG window passes through a dual-pathway convolutional encoder --- a global pathway of convolutional blocks collapsed by learnable attention pooling, and a local pathway preserving six coarse temporal bins --- and a fully-connected head, giving a 256-dimensional per-trial feature. An optional four-layer causal transformer then operates over the sequence of features within a recording run. Separately, the presented word is mapped to a hidden state of Llama-3.1-8B, taken either at layer 0 (non-contextual: the input embedding, which depends only on word identity) or at layer 20 (contextual: dependent on all preceding text), and projected linearly. The two pathways meet only in a shared 256-dimensional L2-normalised space, where they are trained with the symmetric CLIP contrastive objective; the target pathway never sees the EEG. The presence of the transformer and the choice of embedding layer (0 or 20) define the four configurations compared throughout the Results. \textbf{(d)} How scores are computed. Within a fixed pool of 512 validation trials, the model's EEG prediction is scored against every candidate target, softmaxed, and probability mass is summed across trials sharing a core word (case-folded, punctuation stripped) before the rank of the true word is read. The metric reported throughout the present paper is the resulting top-10 accuracy \emph{minus} an empirical permutation baseline, so chance is always zero.}
\end{figure}

\section{Methods}
\label{sec:methods}

\subsection{Participants and data collected}
\label{sec:participants}

One contributor has completed 393 recording runs, each with different text. The text read by the participant was drawn from a corpus of fictional books. In total, the participant contributed 240{,}141 word presentations, corresponding to 48.7\,h of on-task recording. The number of trials passing the amplitude criterion described in Section~\ref{sec:preprocessing} depends on the analysis window (roughly 89--93\%) and is therefore a property of the analysis configuration rather than of the dataset (Supplementary~\ref{sec:s8}).

The present report analyses data from the aforementioned deep subject. Additionally, we have collected silent reading data from 60 participants across one or two sessions each (data collection ongoing). In this first report, we deliberately focus on fundamental questions (can language information be decoded non-invasively at all?) that are best addressed with a large single-subject dataset. Cross-subject model training introduces additional complexity that will be addressed separately in a future report. Here, the goal is generalisation to unseen EEG trials and unseen text (but not unseen words).

\subsection{Stimuli and task}
\label{sec:stimuli}

Text was drawn from fictional books (Sherlock Holmes stories), chosen to be enjoyable, engaging, yet not too cognitively demanding. We think that in the context of language decoding, engaging stimulus material is not just a nice-to-have, but essential, to facilitate subject compliance and continual deep semantic processing of the stimuli. Words were presented one at a time at the centre of a black screen (rapid serial visual presentation). Each run contained approximately 600 words; the exact count varies slightly because text was cut at paragraph boundaries. Within a session, the narrative progressed continuously from run to run, so that the text of run $n{+}1$ continues the text of run $n$.

To decorrelate word identity from low-level visual form, typography was randomised independently on every trial: 7 colours, 37 font families, 3 font sizes, and continuously varying character spacing. This directly implements the recommendation of \citet{csaky2025towards}, who noted that using a single fixed rendering per word facilitates visual form as a confound in silent-reading decoding. Each word remained on screen for 500--900\,ms, scaled by word length (character count) plus up to 100\,ms of random jitter. Specifically, words with up to nine characters were shown for 500\,ms. Words with ten characters or more were shown longer (10 characters: 600\,ms, 11 characters: 700\,ms, 12 characters: 800\,ms, 13 or more characters: 900\,ms). At the beginning of each run, a black screen was shown for 3 seconds. After every 100 words, there was a break (black screen) for 6 seconds.

Two protocol changes occurred early in data collection and are treated as nuisance factors in the analyses:

\begin{itemize}
  \item \textbf{Attention task.} The first 184 runs used a one-back task in which the participant pressed a button when a word repeated. This was replaced by four multiple-choice comprehension questions at the end of each run, which we judged a better guarantee of sustained attention. In all analyses reported here, target events (the second occurrence of a repeated word) and behavioural false alarms are excluded at the trial level, so repeated words never enter the model.
  \item \textbf{Inter-stimulus interval (ISI).} Words were initially separated by a 100\,ms blank screen. Participant feedback during pilot experiments indicated that most participants found text comprehension difficult at ISI $=100$\,ms, possibly resulting from attentional masking, so the ISI was set to 0\,ms (no blank screen between words) and the affected pilot participants were excluded. The deep participant (on whose data the present report is based) did not report this difficulty, possibly due to genuine inter-individual variability, or caused by a training effect, so their first 171 runs, recorded at ISI $=100$\,ms, are retained. All subsequent runs used ISI $=0$\,ms. ISI was therefore included as an explicit factor in the first model training sweep (Section~\ref{sec:experiments}).
\end{itemize}

\subsection{EEG acquisition}
\label{sec:acquisition}

EEG was recorded from 19 scalp electrodes using a dry-electrode system, at a sampling rate of 600\,Hz. For details, see Supplementary~\ref{sec:s0}.

\subsection{Preprocessing}
\label{sec:preprocessing}

Preprocessing was deliberately minimal, following the finding of \citet{defossez2023decoding} that end-to-end architectures gain little from elaborate M/EEG artefact pipelines. Per recording run, each channel was linearly detrended, band-stop filtered around the mains frequency (50\,Hz $\pm1$\,Hz, 4th-order Butterworth), and band-pass filtered between 1 and 80\,Hz (4th-order Butterworth).

Trials were then extracted as fixed-length windows. The window length has to be fixed, even though the trial duration was not. Because presentation duration scales with word length, variable-length epochs would let the decoder read word length directly off the epoch, a non-neural cue that would inflate accuracy. We tested the effect of locking the EEG time window to stimulus onset, stimulus offset, or the centre of the stimulus interval. Moreover, window length was varied across sweeps (0.5, 0.7, 0.9\,s). Optionally, a random temporal shift of up to $\pm50$ or $\pm100$\,ms was applied to the analysis window during training only, as a data-augmentation strategy.

Artefact rejection was a single amplitude criterion: a trial was excluded if any sample within its (augmentation-extended) window exceeded 100\,\muV{} in absolute value. Amplitude checking was performed over the full extended segment so that temporal-shift augmentation can never bring a rejected artefact into view. Included trials were scaled by dividing by 100, so that the working unit is 100\,\muV{} and values lie mainly in $[-1, 1]$, an approach also adopted by recent EEG foundation models \citep{jiang2024large}. An alternative normalisation that additionally subtracted a pre-stimulus baseline was tested in the first sweep. Excluded trials (artefacts, target events, false alarms) are retained as placeholders in the trial sequence so that positional alignment with the language-model targets is preserved, but are masked out of the loss and of all evaluation (Section~\ref{sec:model}).

\subsection{Target representations}
\label{sec:targets}

Each word was represented by a hidden state extracted from a pretrained Llama-3.1-8B model (\url{https://huggingface.co/meta-llama/Llama-3.1-8B}; \citealp{grattafiori2024llama}). The full text of a run was tokenised, a forward pass was run, and for each word the hidden state of the last sub-word token of its ``core'' form (leading and trailing punctuation stripped) was taken, at a specified layer. For example, consider a trial in which a subject read the word ``corkscrew.'' (including a full stop, because the word appeared at the end of a sentence). To retrieve the target embedding for this word, punctuation was stripped, resulting in the core word ``corkscrew'', which the Llama-3.1 tokenizer represents as three tokens: ``\.{G}c'', ``orks'', and ``crew''. In this case, the chosen target embedding is that of the last sub-token, i.e.\ ``crew''. Because the model is causal, a hidden state at any layer above the input embedding incorporates all preceding words.

Two layers were contrasted throughout:

\begin{itemize}
  \item \textbf{Layer 0 --- non-contextual.} The input embedding. Depends only on the word's identity, not on the preceding text. A decoder trained against layer-0 targets cannot profit from tracking the narrative, because the target carries no narrative information. (In this case, for multi-token words, only the last sub-word token determines the embedding.)
  \item \textbf{Layer 20 --- contextual.} A mid-depth hidden state, semantically rich and heavily context-dependent. Targets for consecutive words within a run are strongly correlated. A decoder trained against these targets can profit from tracking the discourse state as well as from identifying the current word. (Llama-3.1-8B has 32 layers; we chose layer 20 rather than the final layer's state because the latter presumably represents expectations about the next token to a larger degree than a middle layer.)
\end{itemize}

Target embeddings are 4{,}096-dimensional and were projected linearly (without bias) into a shared, 256-dimensional decoding space.

\subsection{Model and training objective}
\label{sec:model}

The decoder has three components (Figure~\ref{fig:paradigm}c; full specification in Supplementary~\ref{sec:s2}).

\textbf{EEG encoder.} A per-trial one-dimensional convolutional network with two parallel pathways. A \emph{global} pathway applies four convolutional blocks ($48 \rightarrow 96 \rightarrow 192 \rightarrow 384$ channels, kernel sizes $11 \rightarrow 5$, strides $1 \rightarrow 2 \rightarrow 2 \rightarrow 2$) and collapses time with a learnable attention pooling. A \emph{local} pathway applies three stride-1 blocks ($48 \rightarrow 96 \rightarrow 192$ channels) and pools into six temporal bins, preserving coarse within-window timing. The two pathways are concatenated and passed through a fully-connected head to a 256-dimensional per-trial feature vector. All blocks use ELU activations, batch normalisation and 20\% dropout.

\textbf{Sequence model (optional).} A four-layer causal transformer (256-dimensional, 8 heads, 1{,}024-dimensional feed-forward, 40\% dropout) over the sequence of per-trial features within a recording run. The causal mask mirrors the causal attention of the language model that generated the targets: position $i$ attends only to positions $j \le i$. Excluded trials are replaced by a learnable \texttt{[MASK]} token, so that the sequence stays aligned with the target sequence and the model is informed that a word occurred but its EEG is unusable. Whether the transformer is present at all was a swept factor.

\textbf{Projection and objective.} Encoder (or transformer) outputs and projected LLM targets are mapped into a shared 256-dimensional space, L2-normalised, and trained with the symmetric CLIP contrastive loss \citep{radford2021learning} using a learnable temperature initialised at the CLIP default. One batch consists of 4 or 8 complete recording runs, so a run's ${\sim}600$ trials all contribute in-batch negatives. An option to mask \emph{within-run} negatives (on the reasoning that consecutive contextual targets are near-duplicates and make confusing negatives) was tested and is reported as an optimisation factor.

Models were trained for 100 epochs with AdamW \citep{loshchilov2017decoupled}, weight decay 0.1, and a cosine learning-rate schedule with 5\% warm-up. Checkpoints were selected on the within-run retrieval gain (Section~\ref{sec:metrics}), not on validation loss (because the contrastive loss can be confounded by an optionally learnable temperature parameter).

\subsection{Train/validation split}
\label{sec:split}

The present analysis is based on data from the deep subject only. However, in preparation for the cross-subject analysis (where participants read partially identical text), we had to ensure that no text leaked between training and validation. We therefore used a content-aware split based on clustering sets of 10-grams. Ca.\ 20\% of runs were assigned to the validation set, and the remaining 80\% were assigned to the training set. Consequently, no text chunk can appear on both sides of the split. For the data-scaling analysis, the split was computed first, and only training runs were then subsampled, so that the full validation set is scored identically at every data ratio.

\subsection{Evaluation metrics}
\label{sec:metrics}

Accounting for potential confounds in the evaluation required considerable care, because a contrastive decoder trained against contextual targets has several ways to achieve spuriously high decoding accuracy. We report three regimes rather than a single accuracy. All metrics are word-grouped top-10 retrieval within fixed candidate pools of 512 validation trials. Within a pool, the model's EEG prediction for a trial is compared (cosine similarity, scaled by the learned temperature) against every candidate target in the pool. The resulting distribution is softmaxed and probability mass is summed across trials sharing the same word (lower-cased, punctuation-stripped), so that repeated presentations of the same word are treated as one candidate. The rank of the true word is recorded. Metrics are averaged over five independent pool draws. A fixed pool size gives a well-defined and stable reference level.

Crucially, every reported quantity is a gain relative to an empirical permutation baseline, not a raw accuracy. The baseline is obtained by permuting the EEG predictions within the pool (10 permutations) and recomputing the same metric. This matters because word frequency alone confers a large advantage: a common word occupies many candidate slots and accumulates probability mass regardless of the EEG. The permutation baseline absorbs that, so chance is zero for every gain reported below, and a positive value cannot be produced by the word-frequency structure of the text. Because a gain is a difference of two proportions (model accuracy minus baseline accuracy), we quote all gains in percentage points (pp): a within-run gain of 6.7\,pp means the model's top-10 accuracy exceeds its empirical baseline by 6.7 percentage points, and chance is 0\,pp. Percentage points (absolute differences) should not be confused with the relative changes also quoted in the Results (e.g.\ ``$-32$\%''), which are percentages \emph{of} a gain.

\begin{figure}[tp]
  \centering
  \includegraphics[alt={Schematic of the three evaluation regimes with pool construction, permutation nulls, and a worked example},width=\textwidth]{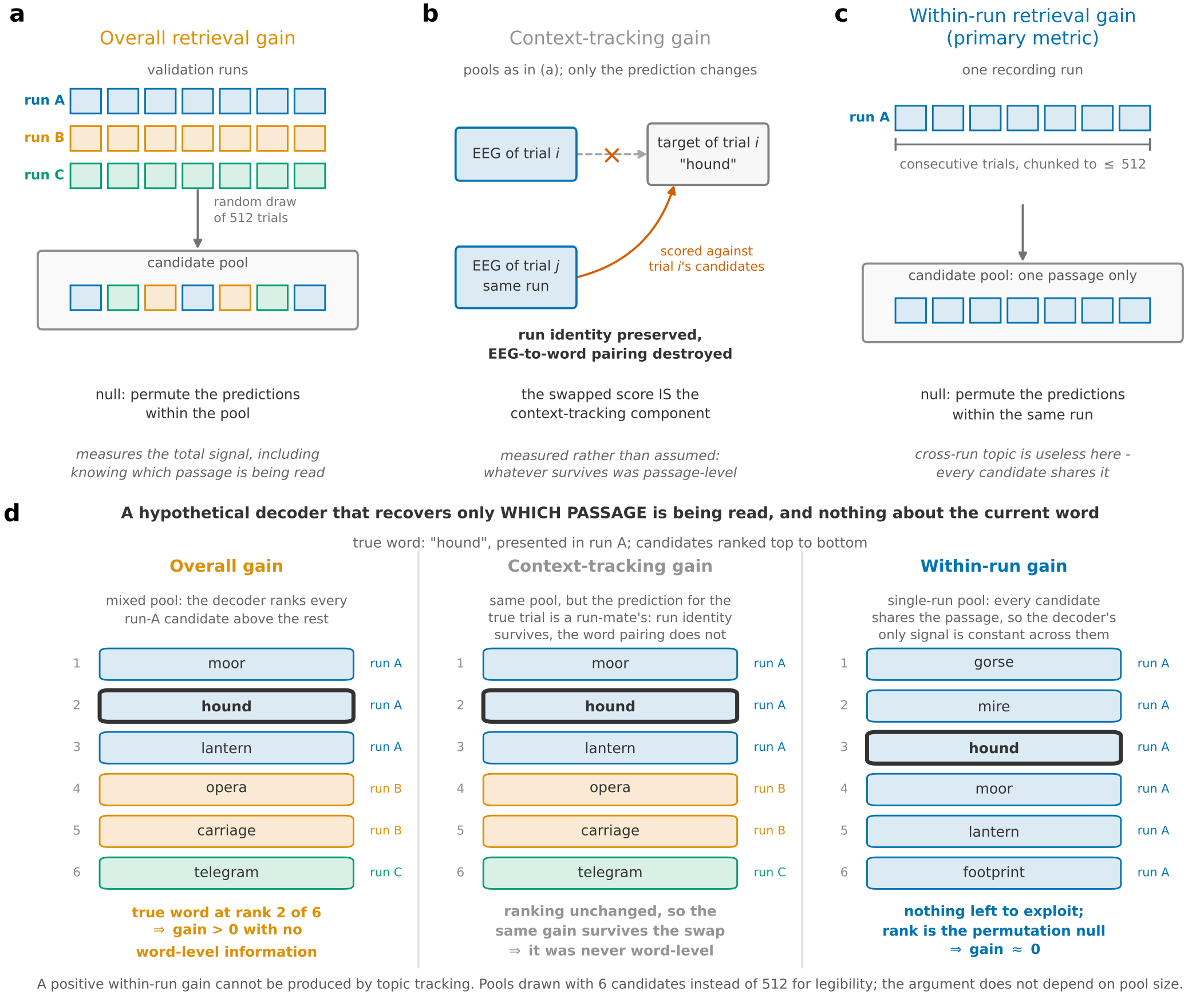}
  \caption{\textbf{The three evaluation regimes, and why the within-run gain is the conservative one.} \textbf{(a-c)} Pool construction and null for each regime; trials are coloured by recording run. \textbf{(a)} \emph{Overall retrieval gain}: pools are random samples of 512 validation trials, so most of a trial's competitors come from other runs and therefore other text passages; the null permutes predictions within the pool. \textbf{(b)} \emph{Context-tracking gain}: the same pools, but each trial's EEG prediction is replaced by that of another trial from the same run. Run identity is preserved while the EEG-to-word pairing is destroyed, so whatever score survives was carried by passage-level context rather than by the current word. Hence, the context component is measured, not assumed. \textbf{(c)} \emph{Within-run retrieval gain}, the primary metric and the checkpoint-selection criterion: pools are consecutive trials from a single run and the permutation null shuffles predictions within that run, so every candidate shares the passage's topic. \textbf{(d)} Worked example for a hypothetical decoder that recovers only which \emph{text passage} is being read and nothing about the current word. In a mixed pool, it ranks every same-run candidate above the rest, and so scores above chance on the overall gain. That score is unchanged by the run-mate swap, which is what makes the swap a measurement of the context-tracking component; and in a single-run pool it has nothing left to exploit and falls back to the permutation null. A positive within-run gain therefore cannot be produced by topic tracking. Pools are drawn with six candidates rather than 512 for legibility; the argument does not depend on pool size.}
  \label{fig:regimes}
\end{figure}

The three evaluation regimes (Figure~\ref{fig:regimes}) are:

\textbf{Overall retrieval gain}. Pools are random samples of validation trials, so a pool typically mixes trials from many recording runs. The baseline in this regime is a permutation within the pool. This is the total decoding signal, and it might be most directly comparable to published segment-identification accuracies. But on its own, the overall retrieval gain is uninterpretable, because in a cross-run pool, the true target's competitors mostly come from \emph{other} runs (with different text), so simply knowing which narrative is being read can rank the target higher. (Run-level EEG variance alone cannot inflate overall retrieval gain. Ranking is $\mathrm{pred}_i \cdot \mathrm{tgt}_j$ across candidates, so a run-identity component in $\mathrm{pred}_i$ only helps if the targets of that run's words also cluster together in the shared space.)

\textbf{Context-tracking gain}. Identical to the overall gain, except that each trial's EEG prediction is replaced by that of another trial from the \emph{same} recording run. This destroys the EEG-to-word pairing. If the gain survives this swap, it was carried by run-level context. In contrast, if it collapses to zero, it required the correct EEG-to-word pairing. This quantity therefore is the context-tracking component of the overall gain, measured rather than assumed.

\textbf{Context-independent gain} $=$ overall gain $-$ context-tracking gain. The portion of the overall gain not explained by cross-run context. (A derived quantity rather than a fourth pooling regime.)

\textbf{Within-run retrieval gain}. This is the primary metric and the checkpoint-selection criterion. Pools consist of consecutive trials from a \emph{single} recording run, and the permutation baseline shuffles predictions \emph{within} that run. Cross-run topic therefore does not help (every candidate shares the run's topic), and the score reflects discrimination among words that occurred close together in the same text. A positive within-run gain cannot be produced by topic tracking, nor by slow EEG drift that merely tags run identity.

There is one caveat regarding the within-run retrieval gain. In conditions using contextual targets (from Llama layer 20), ``discrimination within a run above a within-run shuffle'' still conflates the \emph{current word} with fine-grained \emph{position within the run} (the slow trajectory of the discourse state). The within-run gain is therefore necessary, but not sufficient for lexical decoding. It is cleanest in the non-contextual, no-transformer configuration, where the target carries no positional information. Additionally, instead of relying on the non-contextual configuration alone, we include position probes that measure the positional residual directly (Section~\ref{sec:position-probes}).

\textbf{Word-frequency bins}. Validation words were split into three bins (rare / mid / frequent) by their trial count in the validation set, and all gains were additionally computed per bin, each against its own empirical chance level.

Two caveats apply. First, we have not used a held-out test set or cross-validation; all values are computed on the validation set, so absolute magnitudes are upper bounds and cross-condition orderings are the most reliable output. Second, reported scalars are read at the epoch that maximised the within-run gain, so that metric carries winner's-curse optimism.

\subsection{Position probes}
\label{sec:position-probes}

With a transformer, the position index enters the model through the learned positional embedding. The positional embedding is the only trial-varying input to the model that is not neural. Because of the positional embedding, contextual (layer-20) targets drift systematically along a run. In principle, a positive within-run gain could therefore be earned from position alone. We measure this instead of assuming it, by re-running the trained sequence model with per-trial encoder outputs replaced by the learnable \texttt{[MASK]} token, the same substitution that artefact-rejected trials receive (Section~\ref{sec:model}), so the model remains in distribution.

Three nested information sets are evaluated: (i) \emph{position only}, where every trial's EEG is masked, leaving the model only its positional embedding; (ii) position plus the EEG of \emph{preceding} words, i.e.\ the trial's own EEG is masked; and (iii) the full model. The differences between consecutive levels decompose the within-run gain into a \textbf{position-only} term, a \textbf{preceding-EEG} term, and a \textbf{current-trial} term. The three terms sum to the within-run gain. We define the \textbf{position-corrected within-run gain} as the within-run gain minus the position-only term; it is the portion of the within-run gain that position alone cannot produce.

As a validity check, we performed a control for representation collapse: a model that collapsed to a constant prediction under masking would yield a zero position-only gain regardless of its decoding abilities. This control is described in Supplementary~\ref{sec:s6-caveats}.

The interpretation of the preceding-EEG term deserves emphasis, because it shapes the reading of the results. The model receives no text; apart from the positional embedding, its only input is EEG. To exploit local context at all, the model must first have decoded information about preceding words from their EEG. The preceding-EEG term is therefore neural language decoding at a coarser granularity, not a shortcut, and combining it with the current word's own response is what a deployed decoder should do. Only the position-only term is non-neural, and that is the term the probes exist to bound.

\subsection{Experiments}
\label{sec:experiments}

Three sweeps are reported, comprising 788 model fits in total. All were restricted to the deep participant, and all fits were additionally evaluated with the position probes of Section~\ref{sec:position-probes}.

\textbf{Sweep 1} (576 fits). A fully crossed grid over three scientifically-motivated factors: LLM embedding layer (0, 20), window lock (onset, offset, centre), and presence of the transformer. Additional optimisation factors: temporal-shift data augmentation (0, $\pm100$\,ms), ISI subset (0\,ms only, 100\,ms only, both), normalisation (scaling only vs.\ baseline subtraction and scaling), and two more technical factors (masking of within-run negatives; trainable vs.\ fixed temperature). The EEG window length was fixed at 0.5\,s.

\textbf{Sweep 2} (data scaling; $2 \times 10$ fits). A single configuration swept over the fraction of training runs retained (10\%, 20\%, etc.\ up to 100\% of training runs), separately for the contextual configuration (layer 20 targets, transformer) and the non-contextual configuration (layer 0 targets, no transformer). One fit per ratio.

\textbf{Sweep 3} (192 fits). The headline question was a \emph{channel ablation}: all 19 channels versus a montage with O1, O2, T5 and T6 removed (occipital and posterior-temporal, i.e.\ the visual and ventral-stream sites that drove silent-reading decoding in \citealp{csaky2025towards}). Also crossed: embedding layer (0 or 20), window length (0.5, 0.7, 0.9\,s), transformer presence, and three optimisation factors (temporal-shift data augmentation 0 vs.\ $\pm50$\,ms; batch size 4 vs.\ 8; learning rate 1e-4 vs.\ 1e-5). Window lock was fixed at onset and ISI at both (i.e.\ including all trials irrespective of ISI). Due to the ill-defined spatial resolution of EEG, the channel ablation is no perfect ablation of visual signal, but we still consider its relative effect informative.

Throughout, factors are described as \emph{scientific} (embedding layer, transformer, window lock and length, channel set, data volume) or \emph{optimisation} (batch size, learning rate, normalisation, negative masking, temperature, augmentation).

\section{Results}
\label{sec:results}

All gains reported here are top-10 word-grouped retrieval gains within 512-trial pools, relative to an \emph{empirical permutation baseline}, expressed in \emph{percentage points} (pp; Section~\ref{sec:metrics}), so that \emph{chance is 0\,pp}. Values are means $\pm$ SD across the model fits contributing to each cell, and the number of fits ($n$) is given for every cell in the Supplementary tables.

\subsection{Word-level information is recoverable, and it is not narrative topic tracking}
\label{sec:results-decomposition}

Across the 576 fits of Sweep 1, the within-run retrieval gain, the primary and most conservative metric, averaged $6.7 \pm 2.7$\,pp (median 6.3\,pp; range 1.4 to 15.4\,pp). Every one of the 576 fits produced a positive within-run gain; the same was true of all 192 fits of Sweep 3 (mean $7.8 \pm 2.9$\,pp, range 2.0 to 16.5\,pp) and all 20 fits of the scaling sweep. Because the reported value is read at the epoch that maximised it, a positive minimum is not by itself evidence of signal; the relevant comparison is with the scale of epoch-to-epoch noise on this metric, which is on the order of 0.1 to 0.2\,pp (the pre-training values at epoch 0 lie within $\pm0.2$\,pp of zero in every cell). The observed minimum of 1.4\,pp is an order of magnitude above that, and the median of 6.3\,pp nearly two. Decoding is therefore not marginal or configuration-dependent in the sense of appearing only in favourable cells: it is present everywhere in the explored space, and the sweeps determine its magnitude rather than its existence.

\begin{figure}[tp]
  \centering
  \includegraphics[alt={Grouped bar charts of within-run, overall and context-tracking gains for the four target-type by architecture configurations in Sweeps 1 and 3},width=\textwidth]{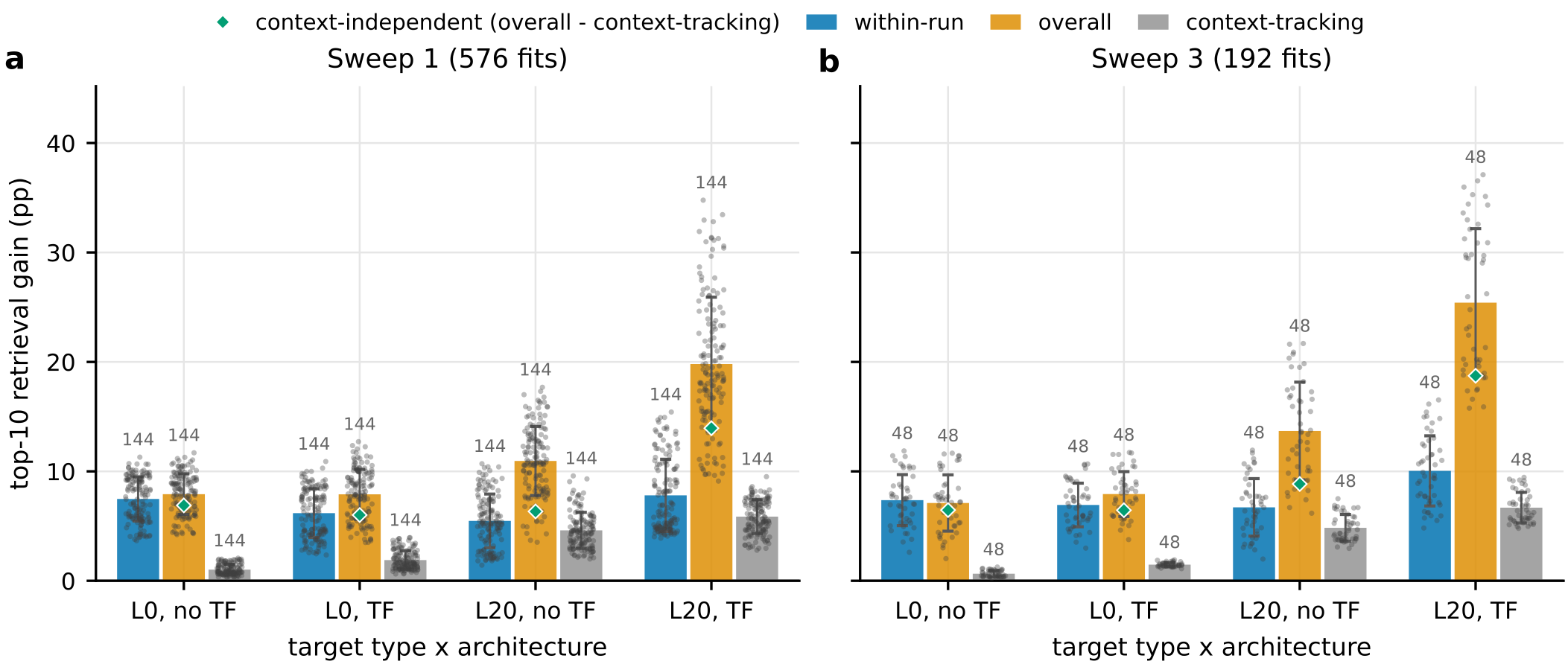}
  \caption{\textbf{Retrieval-gain decomposition by target type and architecture.} Word-level information is recoverable, and it is not restricted to narrative topic tracking. Grouped bars give the three evaluation regimes of Figure~\ref{fig:regimes}, i.e.\ within-run (the primary metric), overall, and context-tracking, for the four crossed configurations of target type $\times$ architecture, in \textbf{(a)} Sweep 1 (576 fits, 144 per configuration) and \textbf{(b)} Sweep 3 (192 fits, 48 per configuration). Configurations are ordered left to right by how much narrative context is available to the fit: neither the target nor the architecture carries it (L0, no TF), one of them does (L0, TF and L20, no TF), or both do (L20, TF). Bars are means over fits $\pm1$ SD, points are individual fits, the number above each bar is $n$, and every quantity is referenced to an empirical permutation baseline, so chance is 0\,pp on the y-axis. The diamond marks the context-independent gain (overall $-$ context-tracking); it is drawn as a derived marker rather than a fourth bar because the three measured gains come from different pooling regimes and therefore do not sum. The decomposition behaves as designed: the context-tracking bar rises with available context (Sweep 1: $1.0 \rightarrow 1.9 \rightarrow 4.6 \rightarrow 5.9$\,pp; Sweep 3: $0.7 \rightarrow 1.5 \rightarrow 4.8 \rightarrow 6.7$\,pp), an ordering predicted by construction. Contextual targets and models (L20, TF) result in by far the largest overall gain, and some of this gain is context tracking.}
  \label{fig:decomposition}
\end{figure}

The decomposition is more informative than the aggregate. Table~\ref{tab:decomposition} and Figure~\ref{fig:decomposition} give the four crossed configurations of embedding layer $\times$ sequence model from Sweep 1.

\begin{table}[tb]
  \centering
  \caption{Retrieval-gain decomposition by target type and architecture (Sweep 1; $n = 144$ fits per cell; mean $\pm$ SD; gains in percentage points, pp).}
  \label{tab:decomposition}
  \small
  \begin{tabular}{lcccc}
    \toprule
    Configuration & \makecell[c]{Within-run\\gain (pp)} & \makecell[c]{Overall\\gain (pp)} & \makecell[c]{Context-tracking\\gain (pp)} & \makecell[c]{Context-independent\\gain (pp)} \\
    \midrule
    \makecell[l]{Non-contextual target,\\no sequence model (L0, no TF)} & $\mathbf{7.5 \pm 2.0}$ & $7.9 \pm 1.9$ & $\mathbf{1.0 \pm 0.5}$ & $6.9 \pm 1.9$ \\
    \addlinespace
    \makecell[l]{Non-contextual target,\\transformer (L0, TF)} & $6.2 \pm 2.2$ & $7.9 \pm 2.3$ & $1.9 \pm 0.8$ & $6.0 \pm 2.2$ \\
    \addlinespace
    \makecell[l]{Contextual target,\\no sequence model (L20, no TF)} & $5.5 \pm 2.5$ & $11.0 \pm 3.2$ & $4.6 \pm 1.6$ & $6.3 \pm 3.0$ \\
    \addlinespace
    \makecell[l]{Contextual target,\\transformer (L20, TF)} & $\mathbf{7.8 \pm 3.3}$ & $\mathbf{19.8 \pm 6.1}$ & $5.9 \pm 1.5$ & $\mathbf{13.9 \pm 5.4}$ \\
    \bottomrule
  \end{tabular}
\end{table}

We find:

\textbf{The decomposition behaves exactly as designed, which validates it.} The context-tracking gain increases monotonically with the amount of contextual information available to the model, from 1.0\,pp when neither the target nor the architecture carries context, through 1.9 and 4.6\,pp when one of them does, to 5.9\,pp when both do. This ordering was predicted by construction and is reproduced in Sweep 3 (0.7, 1.5, 4.8, 6.7\,pp for the same four cells).

\textbf{In the non-contextual configuration, essentially all of the decoding is word-level.} With non-contextual layer-0 targets and no sequence model, neither the target nor the model has any access to narrative context, and the context-tracking gain is correspondingly close to zero (1.0\,pp) while the within-run gain is 7.5\,pp, the second highest of the four cells. Topic tracking cannot contribute here. Short single-trial EEG segments recorded during naturalistic silent reading carry information that discriminates among words presented within the same passage, above a within-passage shuffle of the same EEG, in an open vocabulary.

\textbf{Contextual decoding is much larger overall, and its excess must be decomposed rather than assumed.} The contextual-target-plus-transformer configuration (L20, TF) produces two and a half times the overall gain of the non-contextual configuration (19.8 vs 7.9\,pp) and the single best fits in both sweeps. Of that overall gain, 5.9\,pp (roughly 30\%) is recovered even when a trial's EEG is swapped for a run-mate's, i.e.\ it reflects knowing \emph{which text passage} is being read rather than \emph{which word}. This is a real and arguably useful capability (see Section~\ref{sec:context-position}), but it is not word-level lexical decoding and should not be quoted as such. A possible confounding factor, the transformer's positional embedding acting as a non-neural positional prior on the within-run gain, is measured and discussed in Section~\ref{sec:results-position}.

The interaction between target type and architecture is clean and consistent across sweeps: \textbf{matched} configurations win. Contextual targets require a sequence model to be exploited (contextual embeddings from layer 20 (L20): 5.5\,pp without a transformer $\rightarrow$ 7.8\,pp with one), whereas non-contextual targets are better served without one (layer 0 (L0) embeddings: 7.5\,pp without $\rightarrow$ 6.2\,pp with). A causal transformer trained against a target that carries no context appears to spend capacity modelling sequence structure that the target does not reward, at the cost of per-trial word discrimination.

Selection-optimistic best single fits were a within-run gain of 15.4\,pp (Sweep 1; contextual target, transformer, no augmentation) and 16.5\,pp (Sweep 3; contextual target, transformer, 0.9\,s window, all channels, no augmentation, batch 4, learning rate 1e-4). These are maxima over 576 and 192 fits respectively as well as over epochs, and should be read as upper bounds.

\subsection{The within-run gain is not produced by the positional embedding}
\label{sec:results-position}

\begin{figure}[tp]
  \centering
  \includegraphics[alt={Four-panel figure of position-probe decompositions, dispersion diagnostic, per-frequency-bin corrections, and scaling of probe terms},width=\textwidth]{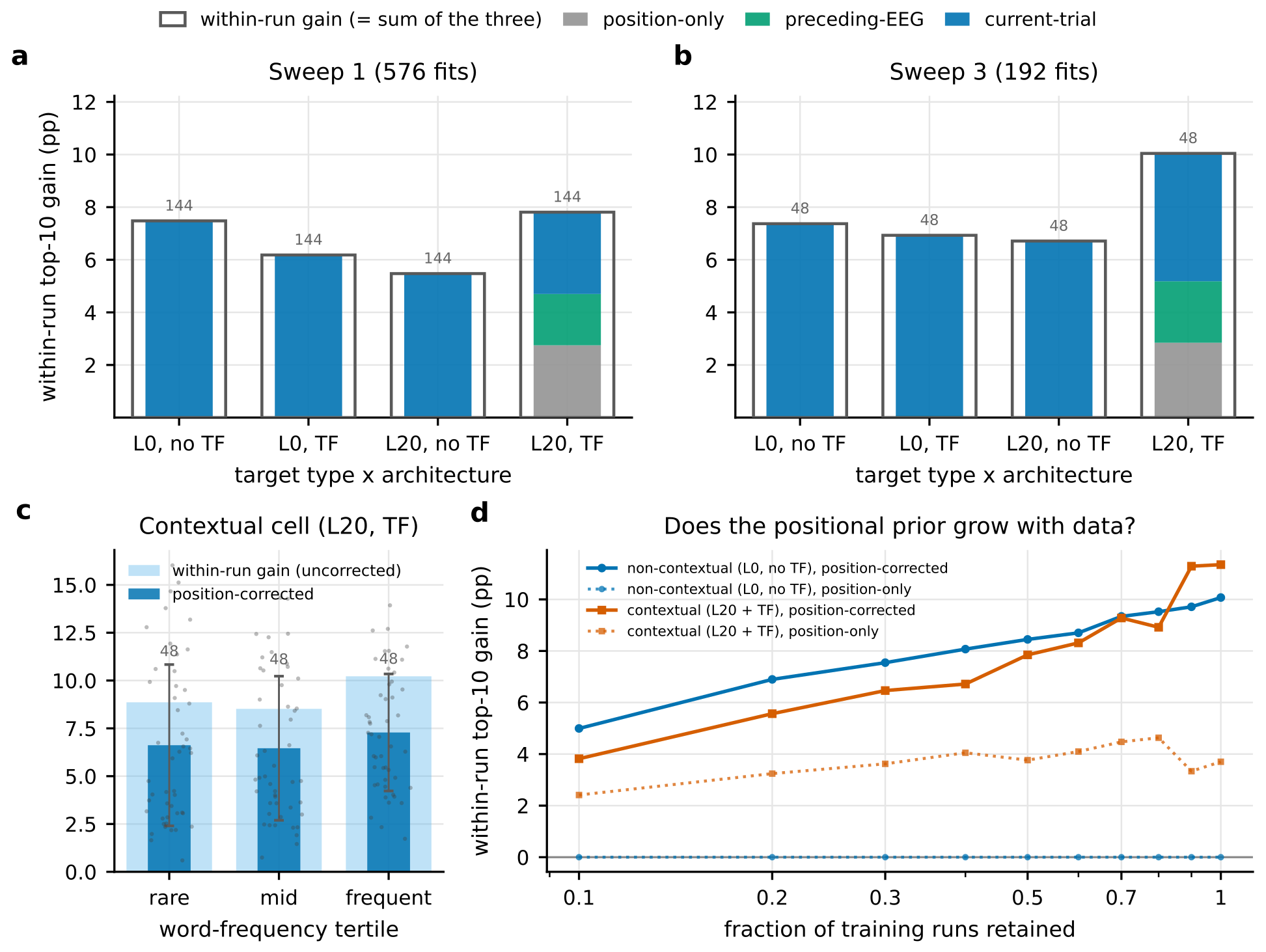}
  \caption{\textbf{Position probes: the within-run gain is not produced by the positional embedding.} The transformer's learned positional embedding is the only trial-varying input to the model that is not neural, and contextual (layer-20) targets drift systematically along a run, so in principle a within-run gain could be earned from position alone. The position probes measure this potential confound by re-running the trained sequence model with per-trial encoder outputs replaced by the learnable \texttt{[MASK]} token (Section~\ref{sec:position-probes}). \textbf{(a)} and \textbf{(b)} The probe results for the four target-type $\times$ architecture configurations, in Sweeps 1 (\textbf{(a)}) and 3 (\textbf{(b)}). The three terms are nested (position only; position $+$ preceding-word EEG; $+$ the current word's own EEG) and sum to the within-run gain. The preceding-EEG segment is a signal, not a nuisance: The model receives no text, so exploiting local context at all requires having decoded preceding words from EEG. Only the position-only segment is non-neural. \textbf{(c)} Raw and position-corrected within-run gain per word-frequency bin. A potential concern is that the decoding performance in some frequency bins might be overwhelmingly driven by exploiting position information. But the results survive position correction in all word frequency bins. \textbf{(d)} Scaling of the position probe terms across a decade of training data (Sweep 2). The position-only term stays flat near zero (L0, no TF) or grows at a lower rate than the corrected component (L20, TF). Bars are means over fits $\pm1$ SD, and chance is zero on every gain axis.}
  \label{fig:probes}
\end{figure}

The within-run gain removes cross-run topic and run-constant drift, allowing us to make conclusions about word-level decoding. However, the within-run gain might still be confounded by a non-neural positional prior (as described in Section~\ref{sec:position-probes}). The position probes (Section~\ref{sec:position-probes}) further decompose the within-run gain into a non-neural positional prior, a preceding-EEG contribution, and a current-trial contribution (Table~\ref{tab:probes}, Figure~\ref{fig:probes}).

\begin{table}[tb]
  \centering
  \caption{Position probes: the within-run gain decomposed into a positional prior, a preceding-EEG contribution, and a current-trial contribution (Sweep 1; $n = 144$ fits per cell; mean $\pm$ SD; gains in percentage points, pp). Unlike Table~\ref{tab:decomposition}, these terms do sum: position-only $+$ preceding-EEG $+$ current-trial $=$ within-run gain, and position-corrected $=$ within-run gain $-$ position-only. Position-only is the only non-neural term. Dispersion is a validity check on the position-only null; a dispersion of zero indicates representation collapse when the model receives masked EEG data and can only use the positional embedding for its prediction. Hence, a near-zero position-only gain is evidence only where dispersion is above zero. (In the transformer-free cells, a dispersion of exactly zero is expected by construction and does not indicate collapse.) For more details, see Supplementary~\ref{sec:s6-caveats}.}
  \label{tab:probes}
  \small
  \setlength{\tabcolsep}{4.5pt}
  \begin{tabular}{lcccccc}
    \toprule
    Configuration & \makecell[c]{Within-run\\gain (pp)} & \makecell[c]{Position-\\only (pp)} & \makecell[c]{Preceding-\\EEG (pp)} & \makecell[c]{Current-\\trial (pp)} & \makecell[c]{Position-\\corrected (pp)} & Dispersion \\
    \midrule
    L0, no TF & $7.5 \pm 2.0$ & $0.0 \pm 0.0$ & $0.0 \pm 0.0$ & $7.5 \pm 2.0$ & $7.5 \pm 2.0$ & $0.000 \pm 0.000$ \\
    L0, TF & $6.2 \pm 2.2$ & $0.0 \pm 0.1$ & $0.0 \pm 0.2$ & $6.1 \pm 2.2$ & $6.1 \pm 2.2$ & $0.167 \pm 0.041$ \\
    L20, no TF & $5.5 \pm 2.5$ & $0.0 \pm 0.0$ & $0.0 \pm 0.0$ & $5.5 \pm 2.5$ & $5.5 \pm 2.5$ & $0.000 \pm 0.000$ \\
    L20, TF & $7.8 \pm 3.3$ & $2.7 \pm 0.7$ & $1.9 \pm 0.8$ & $3.1 \pm 2.7$ & $5.1 \pm 3.0$ & $0.091 \pm 0.048$ \\
    \bottomrule
  \end{tabular}
\end{table}

The position probe allows the following conclusions:

\textbf{The contextual model does exploit positional information.} For L20, TF, position alone earns 2.7\,pp of the 7.8\,pp within-run gain, roughly a third of the cell mean. This result demonstrates that when decoding language from neural signals with context-aware models that have access to positional information, we have to account for the possibility of a non-neural shortcut. Without this control, about a third of the within-run gain for the strongest configuration would have been quoted as neural decoding when it is attributable to a non-neural positional prior.

\textbf{The remaining gain is neural, and it is split between the current and the preceding words.} The position-corrected within-run gain of the contextual configuration is 5.1\,pp, of which 3.1\,pp is contributed by the current trial's own EEG and 1.9\,pp by the EEG of preceding trials (for more details, see Supplementary~\ref{sec:s6-caveats}). The preceding-EEG term is not a confound: the model receives no text, so exploiting local context at all requires having decoded preceding words from their EEG. It is neural language decoding at a coarser granularity, and combining it with the current word's own response is what a deployed decoder should do.

\textbf{After position correction, the contextual and non-contextual configurations decode the current word comparably.} The position-corrected gain of L20, TF (5.1\,pp) does not exceed the non-contextual cell's 7.5\,pp in Sweep 1; in Sweep 3 the two are at parity (7.2 vs 7.4\,pp). Sweep 3 reproduces the whole pattern (L20, TF: within-run 10.0\,pp $=$ 2.9 position-only $+$ 2.3 preceding-EEG $+$ 4.9 current-trial; Supplementary~\ref{sec:s6}). The contextual configuration's real advantages lie elsewhere: in the overall and context-independent gains (Table~\ref{tab:decomposition}), in the rare and mid-frequency vocabulary (Section~\ref{sec:results-frequency}), and in its scaling behaviour (Section~\ref{sec:results-scaling}).

\textbf{The position probe is internally consistent.} In the transformer-free cells, both probes return exactly zero, as they are expected to by construction, and the three terms sum to the within-run gain. In the one cell where a transformer is present but the target carries no context (L0, TF), the position-only term is zero, despite a clearly positive dispersion (0.17): the probe produces position-dependent output and finds essentially nothing where there is nothing to find.

Throughout the remainder of the paper, within-run gains of transformer configurations are read against their position-corrected values, and the position-corrected value is the one we treat as quotable lexical decoding.

\subsection{The signal extends beyond frequent words}
\label{sec:results-frequency}

If the decoder were exploiting only function words and other high-frequency items, the frequency-binned profile would show a gain confined to the frequent bins, but it does not (Table~\ref{tab:freqbins}).

\begin{table}[tb]
  \centering
  \caption{Within-run retrieval gain by word-frequency bin (Sweep 1; $n = 144$ per cell; gains in percentage points, pp).}
  \label{tab:freqbins}
  \small
  \begin{tabular}{lccc}
    \toprule
    Configuration & Rare & Mid & Frequent \\
    \midrule
    L0, no TF & $3.2 \pm 1.5$ & $3.2 \pm 1.4$ & $8.1 \pm 2.2$ \\
    L0, TF & $2.4 \pm 1.5$ & $2.4 \pm 1.4$ & $6.7 \pm 2.4$ \\
    L20, no TF & $3.5 \pm 2.0$ & $3.6 \pm 2.1$ & $5.8 \pm 2.5$ \\
    L20, TF & $6.0 \pm 3.5$ & $6.0 \pm 3.3$ & $8.1 \pm 3.3$ \\
    \bottomrule
  \end{tabular}
\end{table}

The frequent bin does carry the largest gain in every configuration, which is expected. Word frequency confers two advantages: frequent words appear more often in training, so the encoder learns a better mapping for them, and they are represented by more validation trials, so both the model's estimate and the metric's per-bin statistics are better resolved. But the rare and mid bins are also positive throughout. Sweep 3 reproduces this (L0, no TF: rare 3.6, mid 3.8, frequent 7.8\,pp; L20, TF: rare 8.9, mid 8.5, frequent 10.2\,pp).

The gap between the bins narrows in the contextual-plus-transformer configuration: the rare-to-frequent ratio rises from 0.40 in the non-contextual cell to 0.74 in the contextual one (Sweep 1), and from 0.46 to 0.87 in Sweep 3. Two readings are compatible with this, and they are not mutually exclusive. First, on the output side, contextual targets may genuinely improve lexical decoding of rare words. Alternatively, the narrowing may reflect the utility of contextual information at the input side. Rare content words are the most topically distinctive items in a passage, and might therefore both be most indicative of the local discourse state and most predictable from it. A model tracking the semantic trajectory of a run might use correctly decoded information on the topic of the current text passage, in addition to the words' own neural response, to rank them well. The position probes constrain a third reading: the narrowing is not a positional artefact. After subtracting the per-bin position-only term, the rare-to-frequent ratio in the contextual cell remains 0.70 in Sweep 1 (3.7/5.3\,pp) and 0.91 in Sweep 3 (6.6/7.3\,pp), against 0.40 and 0.46 in the non-contextual cell (Supplementary~\ref{sec:s6}).

\begin{figure}[tp]
  \centering
  \includegraphics[alt={Bar charts of within-run and overall gains split by rare, mid and frequent word bins, plus rare-to-frequent gain ratios},width=\textwidth]{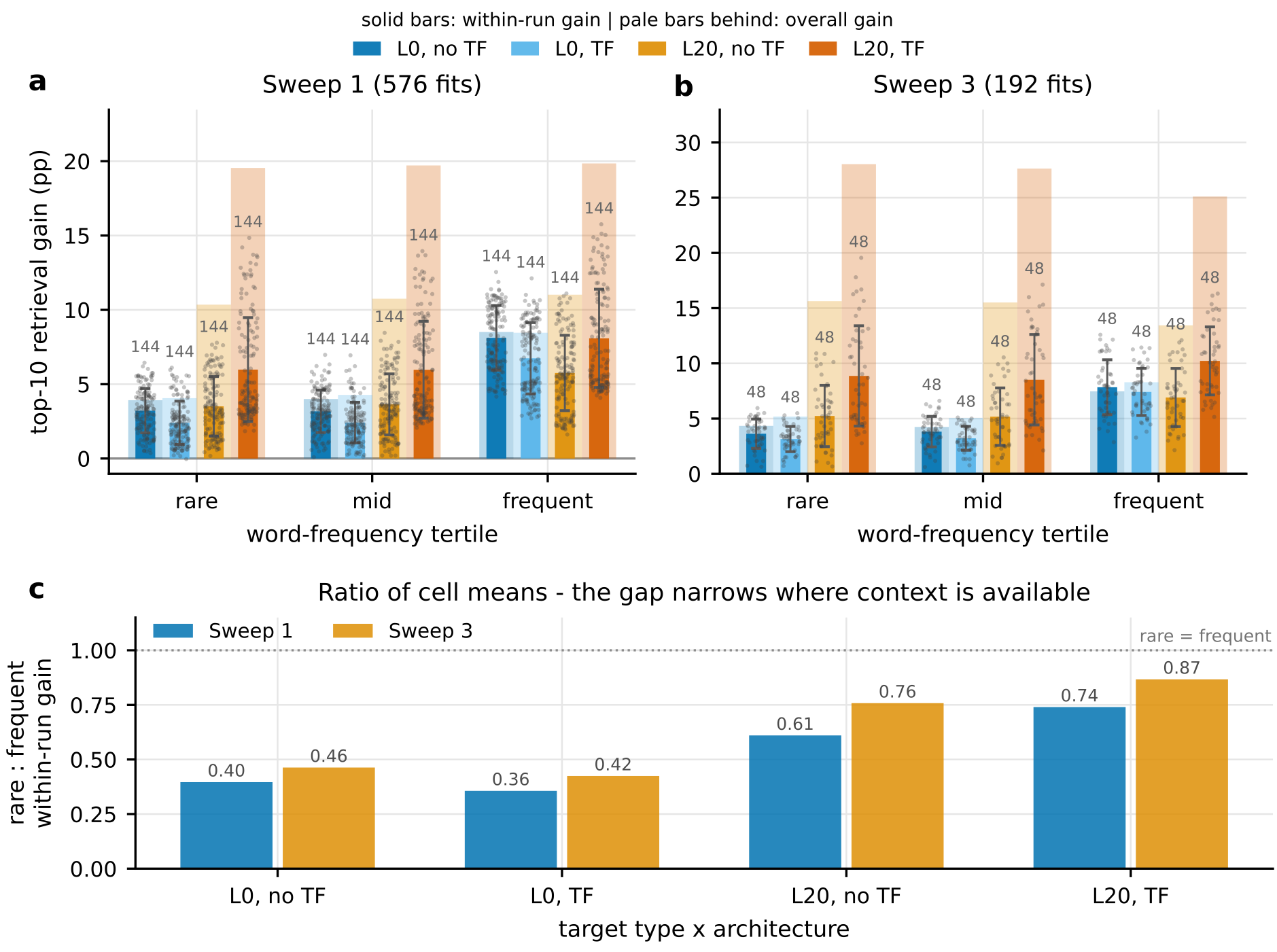}
  \caption{\textbf{Word-frequency profile.} The decodable signal is not confined to frequent words. \textbf{(a, b)} Within-run retrieval gain per word-frequency bin. Validation words were split into rare, mid and frequent thirds by their trial count in the validation set, each scored against its own empirical permutation baseline, for each of the four target-type $\times$ architecture configurations, in \textbf{(a)} Sweep 1 (144 fits per configuration) and \textbf{(b)} Sweep 3 (48 per configuration). The solid inner bar is the within-run gain (the primary metric, word-level decoding); the wider, paler bar behind it is the overall gain from random cross-run pools (including a context-tracking component). Bars are means over fits $\pm1$ SD, points are individual fits, and chance is zero for every bar. The frequent bin carries the largest gain in every configuration, as expected; frequent words are seen more often in training, and are better resolved by the metric (more validation trials). But the rare and mid bins are also positive throughout. A gain confined to function words would appear here as a rare and mid bin at zero, and it does not. \textbf{(c)} Ratio of the rare-bin to the frequent-bin within-run gain per configuration, computed from the cell means for both sweeps. The dotted line at 1.0 marks where the rare bin would decode as well as the frequent one. The gap narrows where narrative context is available, from 0.40 in the purely non-contextual cell to 0.74 in the contextual-plus-transformer cell in Sweep 1, and from 0.46 to 0.87 in Sweep 3. Contextual targets may genuinely improve decoding of rare content words, and / or a model tracking the discourse state may have an advantage when ranking the topically distinctive rare words well, compared to a model that has no context and can only decode them from their own neural response.}
  \label{fig:freqbins}
\end{figure}

\subsection{Decoding scales log-linearly with training data and shows no saturation}
\label{sec:results-scaling}

\begin{figure}[tp]
  \centering
  \includegraphics[alt={Log-linear scaling curves of decoding gain against training-data volume for contextual and non-contextual configurations, overall and per frequency bin},width=\textwidth]{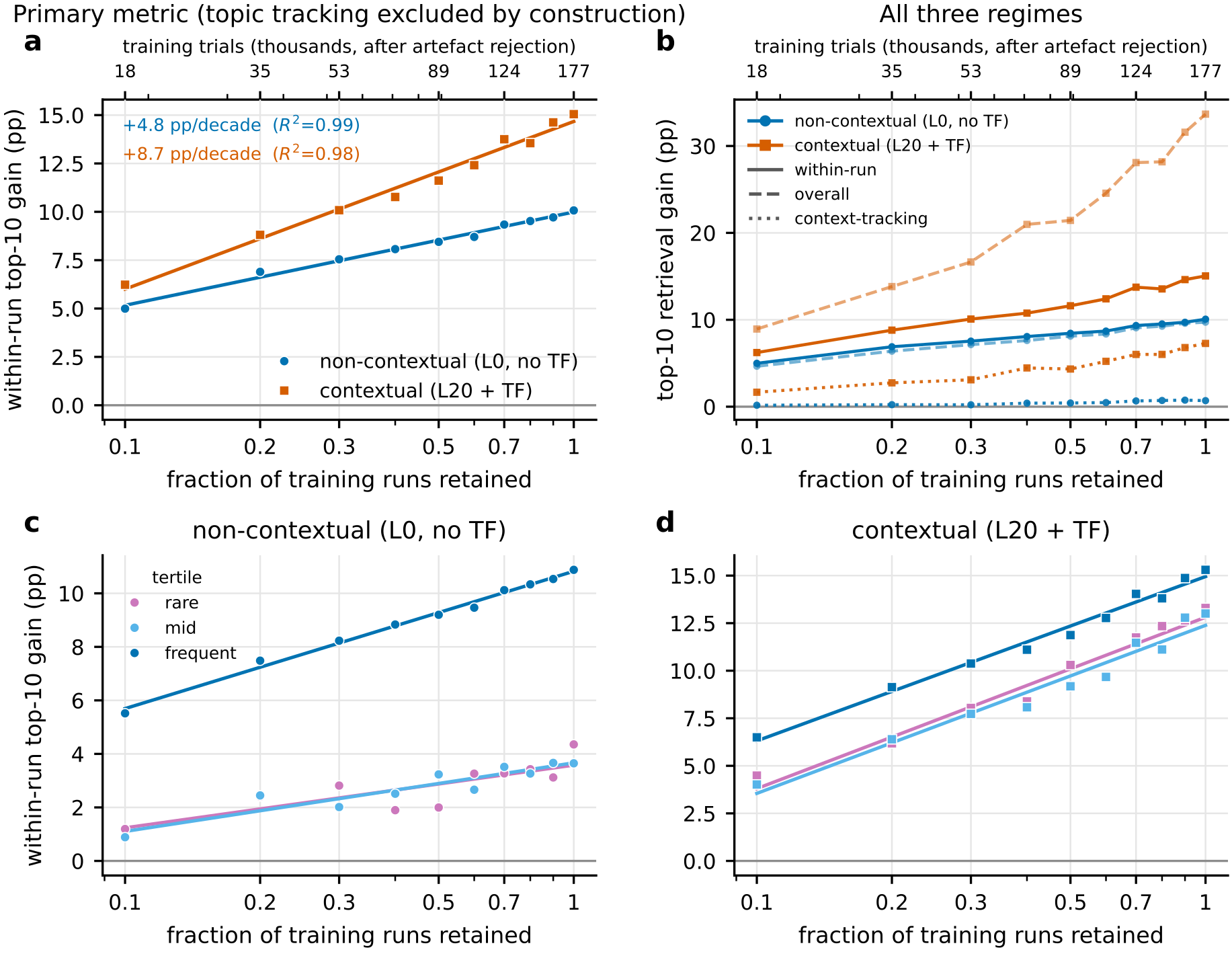}
  \caption{\textbf{Data scaling: decoding grows log-linearly with training-data volume and shows no saturation at ${\sim}49$\,h.} Sweep 2, ten fits per configuration, using 10\%, 20\%, \ldots\ up to 100\% of the training data subset. The lower and upper x-axes show the same information on different metrics (ratio of retained training runs and number of training trials after artefact rejection, respectively). The same, full validation set is used at every ratio. \textbf{(a)} Within-run gain against training volume, with log-linear fit annotated, separately for the most contextual (L20, TF) and least contextual (L0, no TF) conditions. Both configurations roughly double their gain across one decade of data. \textbf{(b)} All three evaluation regimes for most and least contextual configurations (colour $=$ configuration, line style $=$ evaluation regime). Importantly, in the non-contextual condition, the context-tracking trace stays flat against zero ($0.2 \rightarrow 0.7$\,pp, $+0.6$\,pp/decade) while its within-run trace climbs, so the additional data buys word-level decoding rather than progressively better passage recognition. In the contextual condition, the overall gain grows fastest of all ($+24.5$\,pp/decade) but its context-tracking component grows with it ($+5.6$\,pp/decade). \textbf{(c, d)} Per-bin scaling, one panel per configuration. \textbf{(c)} Non-contextual, \textbf{(d)} contextual. Growth reaches the rare and mid bins, at 45--51\% of the frequent bin's rate in the non-contextual arm and at or slightly above the frequent bin's rate in the contextual one. Chance is zero on every gain axis.}
  \label{fig:scaling}
\end{figure}

Two configurations were swept across ten training-data ratios, holding the validation set fixed (Table~\ref{tab:scaling}, Figure~\ref{fig:scaling}). At full data, the training set comprised approximately 192{,}100 trials and the fixed validation set approximately 48{,}000 trials, before applying the amplitude-threshold exclusion criterion (ca.\ 93\% of trials survive the criterion in the 0.5\,s, no-augmentation configuration used here; Supplementary~\ref{sec:s8}).

\begin{table}[tb]
  \centering
  \caption{Log-linear scaling fits, $\mathrm{gain} \sim a + b \cdot \log_{10}(\text{training-data ratio})$. Slopes ($b$) are in top-10 percentage points (pp) per decade, with the $R^2$ of the log-linear fit in parentheses; the endpoint columns give the gain at 10\% and at 100\% of training runs, in pp. One fit per ratio. The probe rows decompose the within-run gain (Section~\ref{sec:position-probes}); ``flat'' means the trace had no variance across the range, which is the desired outcome for the position-only probe.}
  \label{tab:scaling}
  \small
  \setlength{\tabcolsep}{4.5pt}
  \begin{tabular}{lcccc}
    \toprule
    & \multicolumn{2}{c}{Contextual (L20 $+$ TF)} & \multicolumn{2}{c}{Non-contextual (L0, no TF)} \\
    \cmidrule(lr){2-3} \cmidrule(lr){4-5}
    Metric & \makecell[c]{Slope, pp/decade\\($R^2$)} & \makecell[c]{10\% $\rightarrow$ 100\%\\(pp)} & \makecell[c]{Slope, pp/decade\\($R^2$)} & \makecell[c]{10\% $\rightarrow$ 100\%\\(pp)} \\
    \midrule
    Within-run gain (primary) & $\mathbf{+8.7}$ (0.98) & $6.2 \rightarrow 15.0$ & $\mathbf{+4.8}$ (0.99) & $5.0 \rightarrow 10.1$ \\
    Overall gain & $+24.5$ (0.96) & $8.9 \rightarrow 33.7$ & $+5.0$ (0.99) & $4.7 \rightarrow 9.7$ \\
    Context-tracking gain & $+5.6$ (0.94) & $1.7 \rightarrow 7.3$ & $+0.6$ (0.85) & $0.2 \rightarrow 0.7$ \\
    Within-run gain, rare bin & $+9.0$ & $4.5 \rightarrow 13.3$ & $+2.3$ & $1.2 \rightarrow 4.4$ \\
    Within-run gain, mid bin & $+8.8$ & $4.0 \rightarrow 13.0$ & $+2.6$ & $0.9 \rightarrow 3.6$ \\
    Within-run gain, frequent bin & $+8.6$ & $6.5 \rightarrow 15.3$ & $+5.1$ & $5.5 \rightarrow 10.9$ \\
    Probe: position-only & $+1.5$ (0.51) & $2.4 \rightarrow 3.7$ & $0.0$ (flat) & $0.0 \rightarrow 0.0$ \\
    Probe: preceding-EEG & $+1.3$ (0.45) & $0.8 \rightarrow 2.5$ & $0.0$ (flat) & $0.0 \rightarrow 0.0$ \\
    Probe: current-trial & $+5.9$ (0.97) & $3.0 \rightarrow 8.8$ & $+4.8$ (0.99) & $5.0 \rightarrow 10.1$ \\
    Probe: position-corrected & $+7.2$ (0.91) & $3.8 \rightarrow 11.3$ & $+4.8$ (0.99) & $5.0 \rightarrow 10.1$ \\
    \bottomrule
  \end{tabular}
\end{table}

Four observations:

\textbf{Scaling is present in the primary metric.} Both configurations roughly double their within-run gain across one decade of training data, and the log-linear fit is close ($R^2 \ge 0.98$). This is the pattern \citet{sato2024scaling} reported for overt-speech EEG decoding, reproduced here for silent reading and for a metric that has topic tracking removed by construction.

\textbf{Scaling reaches the rare and mid bins.} The rare and mid bins grow at least as fast as the frequent bin in the contextual configuration ($+9.0$ and $+8.8$ vs $+8.6$\,pp per decade) and at 45--51\% of the frequent bin's rate in the non-contextual one, so additional data buys additional word decoding. Hence, all frequency bins profit from more data, but the mid and rare bins improve much faster in the contextual condition (L20, TF) than without context (L0, no TF), similar to the ratios observed in Figure~\ref{fig:freqbins}c.

\textbf{In the non-contextual configuration the context-tracking component stays flat and negligible.} It rises from 0.2 to 0.7\,pp across the full decade (slope $+0.6$\,pp/decade). The gains in that configuration cannot be attributed to the model progressively learning to recognise passages.

\textbf{The positional prior does not grow with the result.} In the contextual configuration, the position-only term grows at $+1.5$\,pp/decade (and with the weakest fit in the table, $R^2 = 0.51$) while the position-corrected gain grows at $+7.2$\,pp/decade. In the non-contextual configuration the position-only trace is essentially zero. A confound that explained the scaling would have to scale with it. In contrast, we observe that across a decade of training data, the additional data is buying neural decoding, not a better positional prior.

Critically, \textbf{none of the curves shows saturation}. At ${\sim}49$\,h of EEG data, the model is data-limited.

\subsection{Removing visual and ventral-stream channels reduces but does not abolish decoding}
\label{sec:results-ablation}

\begin{figure}[tp]
  \centering
  \includegraphics[alt={Electrode layout with ablated channels marked, and bar charts of the ablation effect across evaluation regimes, configurations and frequency bins},width=\textwidth]{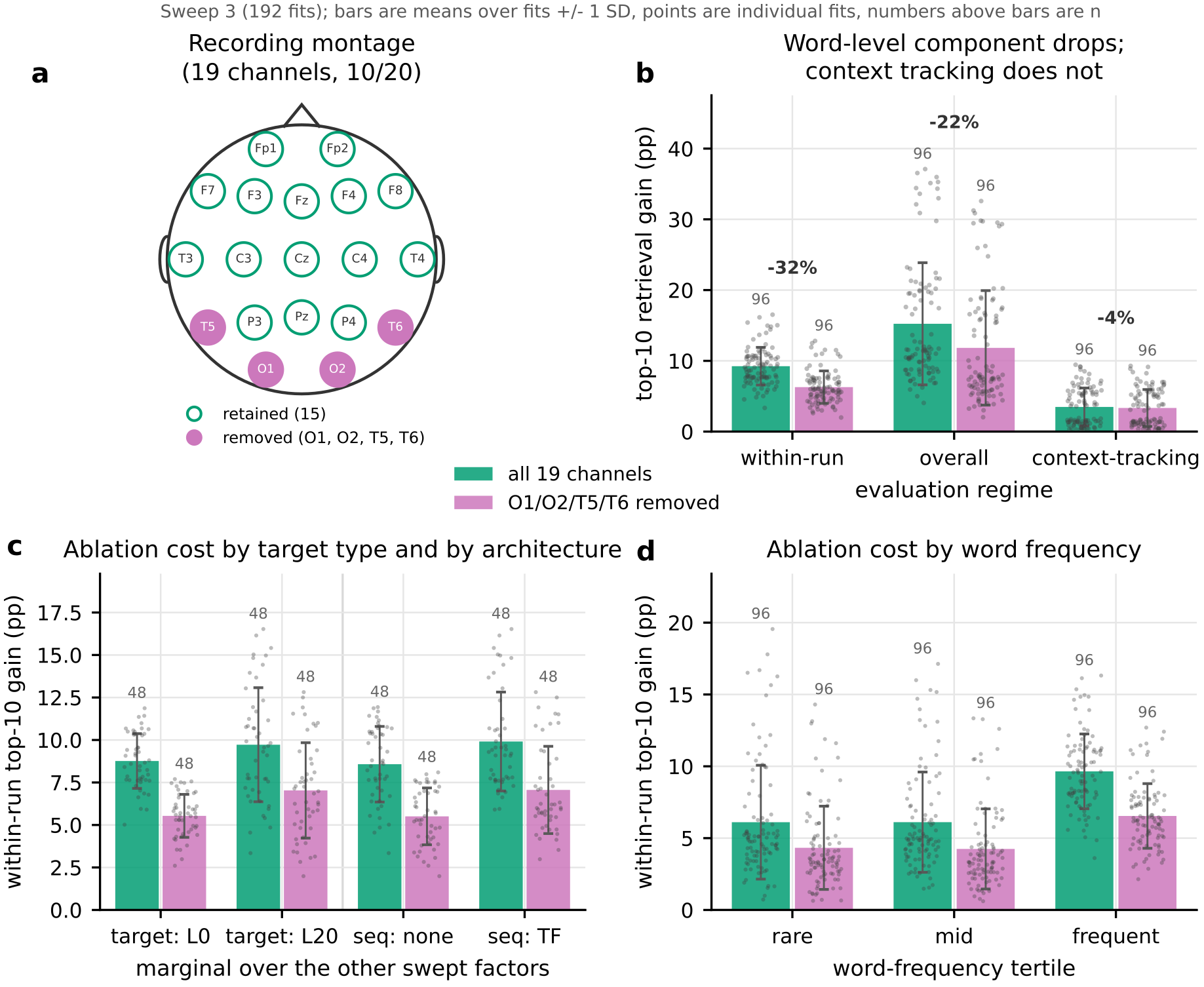}
  \caption{\textbf{Channel ablation: Removing occipital and posterior-temporal electrodes reduces the word-level component by about a third, but leaves passage-level tracking untouched.} All panels are Sweep 3 (192 fits; 96 per channel set, 48 per interaction bar), the only sweep in which the channel set was varied. \textbf{(a)} The 19-electrode 10-20 layout, with the four ablated electrodes (O1, O2, T5, T6; visual cortex and ventral visual stream) marked. \textbf{(b)} The three evaluation regimes with the channel set in the hue, and the relative change of the ablated cell annotated above each pair. The within-run component falls by 32\% ($9.2 \rightarrow 6.3$\,pp) and the overall gain by 22\%, while the context-tracking component moves by only $-4\%$ ($3.5 \rightarrow 3.3$\,pp), well within the dispersion. Whatever supports passage-level tracking is therefore not primarily measured at occipital electrodes, whatever supports word-level discrimination partly is. \textbf{(c)} The ablation cost by target type and by architecture, each bar marginal over the remaining swept factors. The cost is slightly larger in the non-contextual and no-transformer cells ($-37\%$ and $-36\%$) than in the contextual and transformer cells ($-28\%$ and $-29\%$), consistent with the contextual configurations having an additional, non-visual source of signal to fall back on. \textbf{(d)} The ablation cost by word-frequency bin: roughly two thirds of the gain survives in every bin (rare $-29\%$, frequent $-32\%$), so the loss is not concentrated in one part of the vocabulary. Bars are means over fits $\pm1$ SD, points are individual fits, numbers above bars are $n$, and chance is zero throughout. The channel ablation is not a clean localisation; removing 4 of 19 channels removes 21\% of the input, and EEG has ill-defined spatial resolution so occipital sources still project to the remaining electrodes (Section~\ref{sec:results-ablation}).}
  \label{fig:ablation}
\end{figure}

\citet{csaky2025towards} found that silent-reading decoding in EEG, MEG and OPM-MEG was driven by visual processing, with spatial permutation feature importance peaking over occipital sensors and temporal importance peaking near 150\,ms. As an exploratory step, in Sweep 3 we removed electrodes O1, O2, T5 and T6 (over visual cortex and ventral visual stream), leaving 15 of 19 channels (Table~\ref{tab:ablation}, Figure~\ref{fig:ablation}). Because of the ill-defined spatial resolution of EEG, this channel ablation is not a perfect ablation of visual signal.

\begin{table}[tb]
  \centering
  \caption{Effect of removing occipital/posterior-temporal channels (Sweep 3; $n = 96$ per marginal cell, 48 per interaction cell; gains in percentage points, pp).}
  \label{tab:ablation}
  \small
  \begin{tabular}{lccc}
    \toprule
    Contrast & All 19 channels & O1/O2/T5/T6 removed & Relative change \\
    \midrule
    \textbf{Within-run gain (marginal)} & $9.2 \pm 2.7$ & $6.3 \pm 2.3$ & $-32\%$ \\
    Overall gain (marginal) & $15.2 \pm 8.6$ & $11.8 \pm 8.1$ & $-22\%$ \\
    \textbf{Context-tracking gain (marginal)} & $3.5 \pm 2.7$ & $3.3 \pm 2.6$ & $-4\%$ \\
    Position-corrected within-run gain & $8.5 \pm 2.4$ & $5.6 \pm 1.8$ & $-34\%$ \\
    Within-run gain, non-contextual target & $8.8 \pm 1.6$ & $5.5 \pm 1.3$ & $-37\%$ \\
    Within-run gain, contextual target & $9.7 \pm 3.4$ & $7.0 \pm 2.8$ & $-28\%$ \\
    Within-run gain, no transformer & $8.6 \pm 2.2$ & $5.5 \pm 1.7$ & $-36\%$ \\
    Within-run gain, transformer & $9.9 \pm 2.9$ & $7.1 \pm 2.6$ & $-29\%$ \\
    Within-run gain, rare bin & $6.1 \pm 4.0$ & $4.3 \pm 2.9$ & $-29\%$ \\
    Within-run gain, frequent bin & $9.6 \pm 2.6$ & $6.5 \pm 2.3$ & $-32\%$ \\
    \bottomrule
  \end{tabular}
\end{table}

Removing occipital/posterior-temporal channels does affect accuracy, but the decoder does not entirely depend on them. Roughly two thirds of the within-run gain survives their removal, in every configuration and in every frequency bin. The ablation also does not change the ordering of any other factor: longer EEG windows still beat shorter ones, the transformer still helps, no data augmentation (temporal shift) still beats augmentation.

Three features of the pattern are worth noting. First, the \textbf{context-tracking component is almost unaffected} by the ablation ($-4\%$, well within the dispersion), whereas the within-run component drops by roughly a third. Whatever supports passage-level tracking is not primarily measured at occipital electrodes; whatever supports word-level discrimination partly is. Second, the ablation effect is \textbf{slightly stronger in the non-contextual and no-transformer cells} ($-37\%$ and $-36\%$) than in the contextual and transformer cells ($-28\%$ and $-29\%$), which is consistent with the contextual configurations having an additional, non-visual source of signal to fall back on. Third, the \textbf{position-corrected within-run gain falls by 34\%}, closely tracking the raw within-run gain: the ablation cost is borne by the neural component of the signal, not by the positional prior, so the dissociation is not a positional artefact.

Three limits of this ablation experiment are worth noting: (1) Removing 4 of 19 channels removes 21\% of the input data, and performance loss would be expected from that alone. (2) EEG has an ill-defined spatial resolution, so activity generated in occipital and ventral-temporal cortex still projects onto the remaining electrodes. (3) The converse also holds: activity recorded at occipital electrodes is not necessarily entirely visual in nature. Hence, the ablation is an interesting but weak test. A decisive test would be whether a decoder trained on reading transfers to listening, which the planned passive-listening condition is designed to provide.

\subsection{Temporal properties of the informative signal}
\label{sec:results-temporal}

\begin{figure}[tp]
  \centering
  \includegraphics[alt={Bar charts of within-run gain by window lock, window length and training-time jitter, plus a stimulus timeline schematic},width=\textwidth]{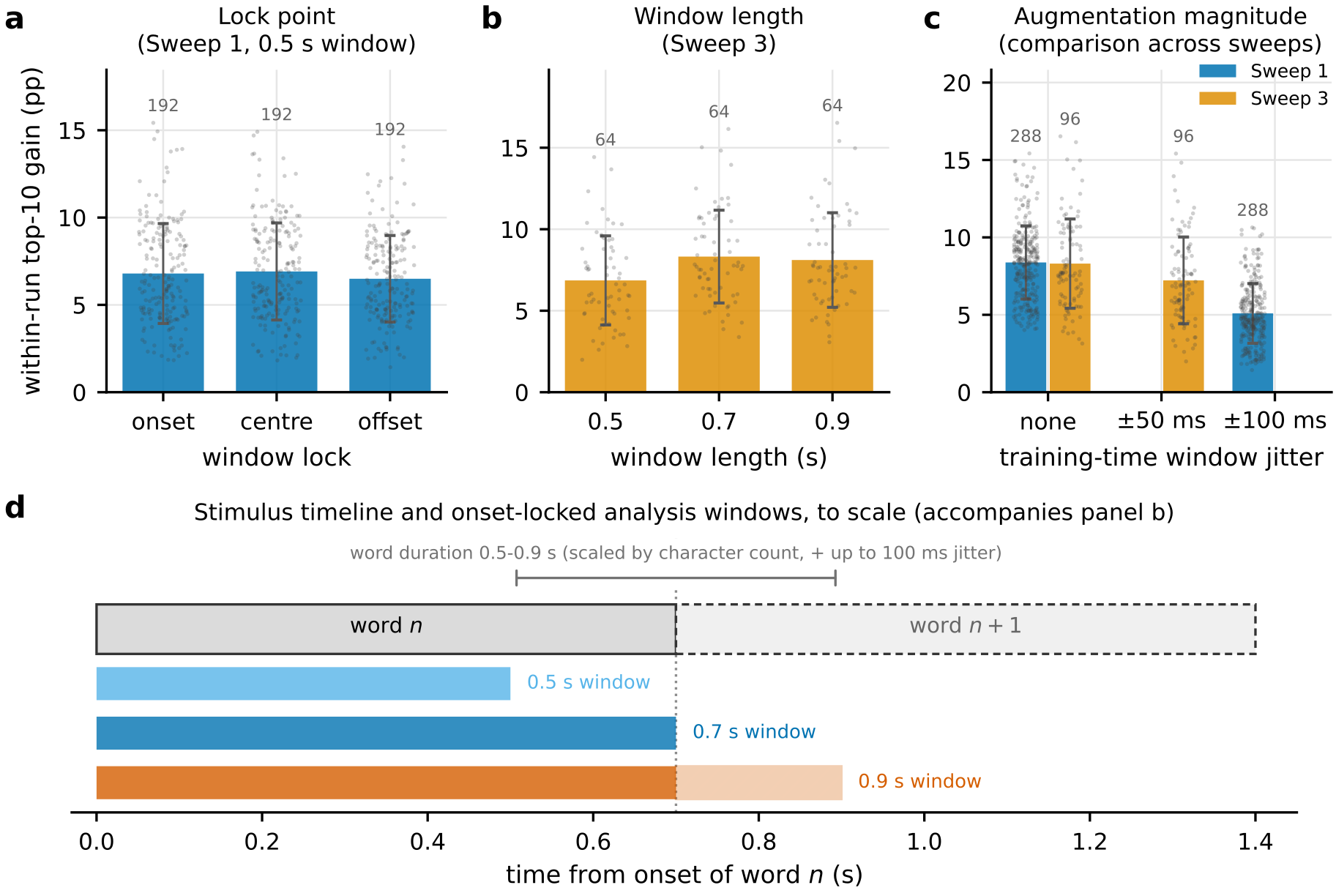}
  \caption{\textbf{Temporal properties of the informative signal.} The lock point barely matters, window length helps up to ${\sim}0.7$\,s, and training-time jitter is harmful. \textbf{(a)} Within-run gain by window lock point (Sweep 1, 0.5\,s windows; 192 fits per level). Onset-, centre- and offset-locked windows fall within 0.4\,pp of each other with overlapping distributions. The decoder is not critically sensitive to window alignment at this granularity. \textbf{(b)} Within-run gain by window length (Sweep 3; 64 fits per level). Extending the window from 0.5\,s to 0.7\,s raises the gain by roughly 20\% ($6.9 \rightarrow 8.3$\,pp); 0.9\,s is no better than 0.7\,s (8.1\,pp). \textbf{(c)} Within-run gain by training-time window jitter ($\pm100$\,ms was crossed in Sweep 1, $\pm50$\,ms in Sweep 3). Jitter is harmful and dose-dependently so: $-13\%$ at $\pm50$\,ms and $-39\%$ at $\pm100$\,ms. This is consistent either with the informative activity being precisely time-locked to word onset, so that jitter smears a phase-locked response, or with the model failing to learn a shift-invariant decoding. In either case, temporal-shift augmentation does not help as a data augmentation strategy in this pipeline. \textbf{(d)} Stimulus timeline and the three onset-locked analysis windows. Depending on the word duration (0.5 to 0.9\,s) and on the window lock (onset, offset, centre), the 0.7\,s and 0.9\,s window lengths can contain the onset of the next word's presentation. The absence of any further improvement from 0.7\,s to 0.9\,s in panel (b) is consistent with the word-specific activity being largely complete within ${\sim}0.7$\,s of onset. Bars in (a), (b), and (c) are means over fits $\pm1$ SD, points are individual fits, and chance is zero on every gain axis.}
  \label{fig:temporal}
\end{figure}

Three analysis-window factors were varied (Table~\ref{tab:temporal}, Figure~\ref{fig:temporal}). First, the window lock, i.e.\ the EEG segments were aligned to the onset of a word, to the centre of the time window during which the word was shown, or to the word offset. Words were shown for 0.5 to 0.9 seconds, depending on the length of the word, but the length of the model's input EEG data was constant (per hyperparameter condition) irrespective of word length (otherwise the model could have learned from a non-neural signal). Second, the window length of the EEG data used as model input. Third, the temporal data augmentation, where the EEG data was shifted in time randomly by up to $\pm50$ or $\pm100$\,ms.

\begin{table}[tb]
  \centering
  \caption{Within-run retrieval gain by analysis-window parameters (gains in percentage points, pp).}
  \label{tab:temporal}
  \small
  \begin{tabular}{llcc}
    \toprule
    Factor & Level & Within-run gain (pp) & Sweep \\
    \midrule
    Window lock & Onset & $6.8 \pm 2.9$ & 1 \\
    & Centre & $6.9 \pm 2.8$ & 1 \\
    & Offset & $6.5 \pm 2.5$ & 1 \\
    \midrule
    Window length & 0.5\,s & $6.9 \pm 2.7$ & 3 \\
    & 0.7\,s & $\mathbf{8.3 \pm 2.8}$ & 3 \\
    & 0.9\,s & $8.1 \pm 2.9$ & 3 \\
    \midrule
    Temporal-shift augmentation & none & $\mathbf{8.4 \pm 2.4}$ & 1 \\
    & $\pm100$\,ms & $5.1 \pm 1.9$ & 1 \\
    & none & $\mathbf{8.3 \pm 2.9}$ & 3 \\
    & $\pm50$\,ms & $7.2 \pm 2.8$ & 3 \\
    \bottomrule
  \end{tabular}
\end{table}

\textbf{Window lock made almost no difference.} Onset-, centre- and offset-locked windows performed within 0.4\,pp of each other, with overlapping distributions. This is perhaps not surprising. With 0.5\,s windows and stimulus durations of 0.5--0.9\,s, the three lock points define mostly overlapping segments, and in a continuous RSVP stream the response to one word overlaps the presentation of the next. The result establishes that the decoder is not critically sensitive to window alignment at this granularity.

\textbf{Longer windows helped, up to a point.} Moving from 0.5\,s to 0.7\,s raised the within-run gain by roughly 20\%, with 0.9\,s no better than 0.7\,s. Because the ISI was zero for most runs, a 0.9\,s window typically extends into the following word's presentation; the absence of further improvement at 0.9\,s is therefore consistent with the informative, word-specific activity being largely complete within ${\sim}0.7$\,s of word onset.

\textbf{Temporal jitter was consistently harmful, and dose-dependently so.} Randomly shifting the analysis window by up to $\pm100$\,ms during training reduced the within-run gain by 39\% ($8.4 \rightarrow 5.1$\,pp); $\pm50$\,ms reduced it by 13\% ($8.3 \rightarrow 7.2$\,pp). This was among the largest effects in Sweep 1 and it was uniform across every other factor. Two readings are compatible with the data and are not mutually exclusive: the informative activity might be time-locked to word onset, so jitter smears a phase-locked response; and / or the model fails to learn a shift-invariant decoding of the signal. The finding parallels \citet{mugler2014direct}, who reported that ECoG phoneme classification degraded sharply once onset alignment was jittered beyond 100\,ms. For present purposes the practical conclusion is straightforward: temporal-shift does not help as a data augmentation method in our current pipeline.

\subsection{Optimisation factors}
\label{sec:results-optimisation}

\begin{figure}[tp]
  \centering
  \includegraphics[alt={Five-column figure of optimisation-factor effects on within-run gain and the corresponding checkpoint-selection epochs},width=\textwidth]{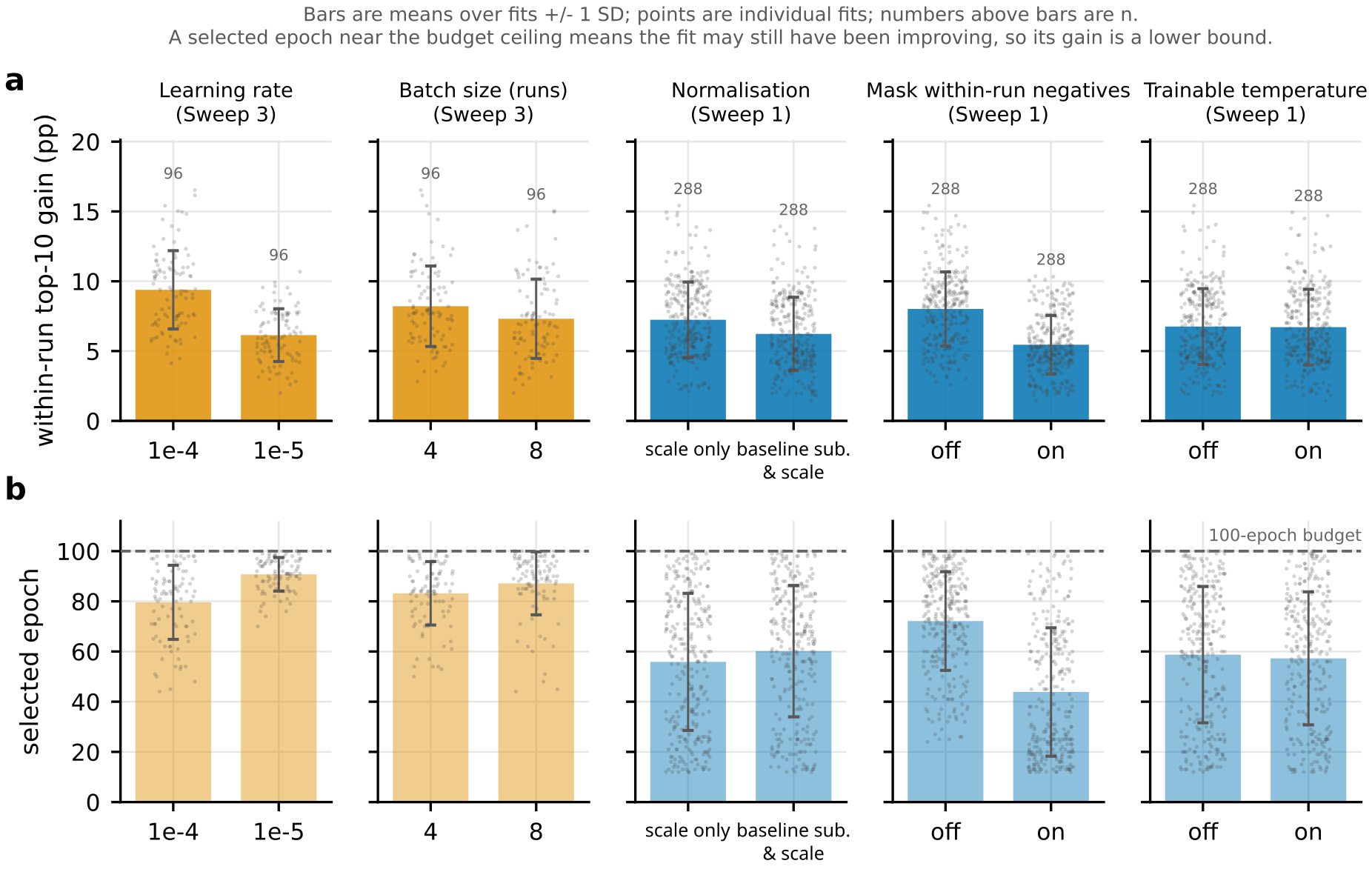}
  \caption{\textbf{Optimisation factors, and how much of each difference is a training-budget artefact rather than an optimum.} Five factors, one column each. \textbf{(a)} Top row: within-run gain per level. \textbf{(b)} Bottom row: the epoch at which that fit's checkpoint was selected, with the fixed 100-epoch budget drawn as a dashed ceiling. A bar sitting against the ceiling identifies a comparison in which the losing level was plausibly still improving when training stopped, so its gain is a lower bound and the difference should not be read as an intrinsic optimum. Sweeps are separate, learning rate and batch size are marginals of Sweep 3, normalisation, within-run negative masking and trainable temperature marginals of Sweep 1, so every comparison is a clean marginal of one fully crossed grid. Bars are means over fits $\pm1$ SD, points are individual fits, numbers above bars are $n$, chance is zero. \emph{Learning rate}: 1e-4 clearly beats 1e-5 (9.4 vs 6.1\,pp), but the 1e-5 fits select their checkpoint at epoch $91 \pm 7$ out of 100 against $80 \pm 15$ for 1e-4, i.e.\ the difference is partly an incomplete budget. \emph{Batch size}: 4 runs beats 8 (8.2 vs 7.3\,pp), with the same caveat ($83 \pm 13$ vs $87 \pm 13$ epochs) plus two structural confounds; doubling the batch halves the number of optimiser steps per epoch, and because a batch item is a complete recording run it also doubles the number of in-batch contrastive negatives. \emph{Normalisation}: simple amplitude scaling beats scaling with pre-stimulus baseline subtraction (7.2 vs 6.2\,pp). \emph{Within-run negative masking}: masking same-run pairs out of the contrastive loss \emph{reduces} the gain ($8.0 \rightarrow 5.5$\,pp), the opposite of its motivation, so discriminating between adjacent words is the useful part of the training signal rather than a source of confusion. \emph{Trainable temperature}: no detectable effect on any gain, which is the expected and desired null. The word-grouped metrics are temperature-dependent, so a null here is evidence they are undistorted.}
  \label{fig:optimisation}
\end{figure}

The optimisation factors were swept to maximise decoding accuracy and to verify that they do not invert the scientific comparisons, and the results are reported briefly (Figure~\ref{fig:optimisation}).

\textbf{Data volume via ISI subsetting.} Training on runs from both ISI conditions gave the highest within-run gain (7.2\,pp), followed by ISI $=0$\,ms alone (6.9\,pp) and ISI $=100$\,ms alone (6.1\,pp). The ordering tracks the number of available recording runs (393, 222 and 171, respectively) rather than any plausible ordering of ISI quality, so we read this as a data-volume effect and not evidence that either ISI condition improves the signal. The two ISI conditions are pooled in all subsequent sweeps.

\textbf{Learning rate and batch size.} A learning rate of 1e-4 substantially outperformed 1e-5 (9.4 vs 6.1\,pp), and a batch of 4 runs outperformed 8 (8.2 vs 7.3\,pp). Both differences are partly training-budget artefacts under a fixed 100-epoch schedule: the 1e-5 fits selected their checkpoint at epoch $91 \pm 7$ (versus $80 \pm 15$ for 1e-4), i.e.\ they might still be improving at the end of training, and doubling the batch halves the number of optimiser steps per epoch. Batch size also changes the contrastive task itself, since a batch is a set of complete recording runs and the number of in-batch negatives scales with it.

\textbf{Normalisation.} Simple amplitude scaling outperformed scaling with pre-stimulus baseline subtraction (7.2 vs 6.2\,pp).

\textbf{Masking of within-run negatives.} Excluding same-run pairs from the contrastive loss, motivated by the near-duplicate nature of consecutive contextual targets, \emph{reduced} the within-run gain, and the penalty was largest in exactly the configuration the manipulation was designed to help (contextual target $+$ transformer: 5.0\,pp masked vs 10.6\,pp unmasked). Discriminating between adjacent words appears to be the useful part of the training signal rather than a source of confusion. Note that masking changes the scale of the loss, so loss values are not comparable across this factor; all rankings here are on gains.

\textbf{Temperature.} Making the CLIP temperature learnable had no detectable effect on any gain. No fit in any sweep exhibited temperature runaway (maximum final temperature 19.3 against a clamp at 100), so the word-grouped metrics (which are temperature-dependent) are not distorted.

\textbf{Control check.} Neither technical factor inverted the ordering of any scientific factor. The full control results are in Supplementary~\ref{sec:s7}.

\section{Discussion}
\label{sec:discussion}

\subsection{Summary}
\label{sec:summary}

From roughly 49\,h of 19-channel EEG recorded from one participant reading continuous narrative prose, an open-vocabulary contrastive decoder recovers information that discriminates among words presented within the same passage. The effect is present in all 788 model fits we ran, survives a within-passage permutation null, extends to mid-frequency and rare words as well as frequent ones, and is partially but not entirely a result of tracking the narrative topic. In the configuration where topic tracking is architecturally impossible, the within-run gain is 7.5\,pp. We have identified positional embeddings as a potential non-neural confound. Position alone accounts for 2.7\,pp of the contextual configuration's 7.8\,pp within-run gain, and the position-corrected remainder is carried by neural signal from the current and the preceding words. Performance grows log-linearly with training data with no sign of saturation, and the growth reaches the rare and mid-frequency bins rather than being confined to frequent words. Removing occipital and posterior-temporal electrodes costs about a third of the within-run gain, but leaves the majority intact, and leaves the context-tracking component essentially untouched.

We regard five of these as substantive results: (1) open-vocabulary word-level decoding from EEG during silent reading of naturalistic text is possible; (2) the decoding can be decomposed into word-level and context-tracking components; (3) the positional prior that a causal sequence model introduces is an important potential confound that can be measured, and the within-run gain is not reducible to it; (4) the system is in a data-limited regime; (5) the signal is not exclusively measured at occipital and posterior-temporal electrodes.

\subsection{Relation to prior non-invasive work}
\label{sec:prior-work}

The most direct comparison is \citet{csaky2025towards}, who also targeted silent reading with EEG and MEG in a densely-sampled participant. They decoded five words at 30--40\% against a 20\% chance level. Our design differs in ways that make a numerical comparison meaningless but a qualitative one informative. Our vocabulary is open and natural (thousands of unique words drawn from continuous prose rather than five clinically-chosen words), our typography is randomised trial-by-trial, our decoder is contrastive against language-model embeddings rather than a linear classifier over covariance features, and our evaluation is retrieval within 512-candidate pools against a permutation baseline. The convergent conclusion is that silent reading is decodable from EEG. Our contribution is that it is decodable at the level of individual words drawn from an open vocabulary, and that the decodable information is not exhausted by the occipital and posterior-temporal electrodes that their permutation feature importance identified as the driver.

In the EEG component of their study, \citet{defossez2023decoding} achieved top-10 segment accuracies of 17.7\% and 25.7\% out of 1{,}842 and 190 candidates respectively, for perceived speech. Our task, metric, candidate-set construction and chance level all differ, so we do not attempt an accuracy comparison. What transfers is the methodological lesson their ablations established and our results reinforce: the contrastive objective and pretrained deep target representations enable non-invasive decoding of natural, open-vocabulary language.

The most relevant comparison for the scaling result is \citet{sato2024scaling}. Their 175\,h single-participant overt-speech EEG dataset reached 48.5\% top-1 and 76.0\% top-10 over 512 candidate segments, and exhibited an unsaturated log-linear relationship between data volume and accuracy, with performance collapsing when subsampled to ${\sim}3$\,h. Our current deep-subject data volume is roughly a quarter of theirs, our unit is a single word rather than a 5\,s multi-word segment, our task is silent reading rather than reading aloud, and our metric is referenced to a permutation baseline rather than to uniform chance. Within those differences, we reproduce the qualitative scaling result on a metric constructed to exclude topic tracking, which their design did not separate out. We also avoid the concern that dominated their discussion: silent reading generates no speech-related electromyographic activity, so the elaborate EMG-decontamination they required (adaptive filtering plus adversarial data augmentation) has no analogue here. We trade that concern for a different one, the visual confound, discussed in Section~\ref{sec:visual-confound}.

Taken together, a coherent picture emerges. Non-invasive language decoding fails when data are scarce and succeeds, partially, when they are abundant. The paradigms that yield abundant, well-time-locked data are perception and overt production. The open question is whether anything learned in those regimes will generalise to inner speech.

\subsection{Context, position, and what counts as decoding}
\label{sec:context-position}

Every high-performing invasive speech neuroprosthesis depends heavily on linguistic priors. \citet{moses2021neuroprosthesis} reported that a language model reduced their median word error rate by 35 percentage points; \citet{metzger2023high} and \citet{card2024accurate} both rely on n-gram language models and large-language-model rescoring to convert noisy phoneme posteriors into accurate text; \citet{kunz2025inner} decode a 125{,}000-word vocabulary in the same way. A decoder that ``knows'' what is being talked about is a decoder whose language-model prior is better conditioned. In a deployed system, a component that tracks discourse state from EEG and a component that discriminates the current word would be complementary.

The reason we separate context tracking and word-level decoding is to gain a clearer understanding of the capabilities of our decoding model. Contextual language-model targets introduce the risk of obtaining large and impressive-looking retrieval scores that contain little word-level information, and the field has to be cautious of this class of problem. In a previous example from the EEG-to-text literature, accuracies that appeared strong turned out to depend on teacher forcing at evaluation and dropped to chance without it (as reviewed in \citealp{sato2024scaling}). Our contextual-plus-transformer configuration produces an overall gain of 19.8\,pp, of which 5.9\,pp is measurably attributable to passage identity.

The position probes extend this control from context to position, and we regard them as a central methodological strength of the present study. With a sequence model, the learned positional embedding is the only trial-varying input to the decoder that is not neural, and contextual targets drift systematically along a run (Section~\ref{sec:position-probes}), so a model can earn an apparent within-run gain merely by knowing \emph{where in the run} a trial occurred. The probes measure this shortcut directly. In our strongest configuration, position alone accounts for 2.7\,pp of the 7.8\,pp within-run gain, roughly a third of what would otherwise have been quoted as word-level decoding. All conclusions about contextual configurations are therefore drawn from position-corrected values. The same probes also isolate the contribution of the \emph{preceding} words' EEG, and this term is not a confound: the model receives no text, so any use of local context requires having decoded the preceding words from their EEG. It is neural language decoding at a coarser granularity, and it mirrors the way the invasive systems mentioned above condition their language-model priors on previously decoded history.

The measured decomposition also sharpens the comparison between contextual and non-contextual configurations. The non-contextual, no-transformer cell serves as an important baseline, because there the positional and contextual routes are absent by construction rather than by measurement. Once the positional prior is subtracted, the contextual configuration's advantage is neural, but it does not lie in a larger current-word signal at the present data volume (position-corrected within-run gain: 5.1 vs 7.5\,pp in Sweep 1, 7.2 vs 7.4\,pp in Sweep 3). It lies instead in the overall and context-independent gains (where decoded context is a legitimate additional discriminator), in the rare and mid-frequency vocabulary, and in a steeper scaling slope ($+7.2$ vs $+4.8$\,pp per decade, position-corrected). We would encourage future studies that align brain data to contextual language-model embeddings to report word-level and contextual components where applicable, and to safeguard against inflated accuracies from non-neural positional-embedding cues.

\subsection{What this does and does not imply for inner speech}
\label{sec:implications-inner-speech}

Previous studies and our results support the view that non-invasive language decoding requires large amounts of training data. Subject compliance and attention cannot be sustained long enough when using slow-paced, unnaturalistic, boring tasks. Instead, silent reading and passive listening can act as proxy tasks to provide the bulk of training data for EEG-to-text decoders. Whether a model trained on such a proxy task can transfer to spontaneous inner monologue remains an open question.

The invasive literature is encouraging on the representational question. \citet{martin2016word} provided one of the first demonstrations that individual imagined words are decodable from direct cortical recordings, with discriminative information in superior temporal gyrus, inferior frontal gyrus and sensorimotor cortex. \citet{wandelt2024representation} found strong shared representations in supramarginal gyrus between internal speech, visually presented word reading, and vocalised speech, and reported that decoders trained on written-cue trials generalised better to internal and vocalised speech than decoders trained on auditory-cue trials, a direct argument that reading-based training data is a reasonable proxy for an inner-speech decoder. \citet{kunz2025inner} showed that inner speech occupies a shared, scaled-down subspace with attempted speech in motor cortex, so that training on attempted speech transfers. \citet{jones2021note} demonstrated zero-shot transfer in the specific direction we care about: deep networks trained on elicited inner speech (covert reading and repeating) decoded self-generated inner speech at an accuracy matching within-task replication, in 7T fMRI. The representational precondition for transfer therefore appears to hold.

What is unresolved is whether it survives the signal-to-noise and spatial-resolution penalty of scalp EEG. \citet{csaky2025towards} could decode silent reading but not inner speech, and the failure of the inner-speech decoders prevented them from even testing transfer. Our reading of that result is not that non-invasive inner-speech decoding is impossible, but that it was attempted with less data per participant than the scaling results suggest is needed, on a five-word vocabulary, with paradigms whose compliance could not be verified. Our strategy is to build a high-volume, high-compliance, well-time-locked proxy corpus first; establish that a decoder trained on it extracts language-related rather than modality-specific information (the future reading/listening comparison); and only then attempt transfer to inner speech. The present results support the first step.

\subsection{The visual-confound question}
\label{sec:visual-confound}

The single most important caveat on any silent-reading result is that the decoder might be reading out visual word-form processing instead of deep semantic language processing. \citet{ling2019visual}, decoding 80 words from EEG during silent reading, found that pairwise discriminability peaked around 170\,ms and correlated with visual and orthographic similarity but not with semantic similarity, demonstrating that EEG word decoding \emph{can} be entirely visual. \citet{csaky2025towards} localised their silent-reading decoding to visual areas with an early temporal peak.

We have taken three measures to address the visual confound. First, typography is randomised on every trial, so word identity is decorrelated from any fixed visual template. This partially removes a specific confound Csaky et al.\ flagged, though it does not remove the systematic visual differences between different letter strings. Second, we compared non-contextual with contextual language-model targets. The layer-20 targets are semantic and context-dependent and bear no systematic relation to orthographic form, so the fact that they support the largest gains is at least consistent with non-visual information contributing. Third, we ablated the occipital and posterior-temporal channels, and roughly two thirds of the within-run gain survived.

None of the three measures is decisive. Volume conduction means occipital sources can reach frontal and central electrodes; a signal that survives occipital-channel removal has not been shown to have a non-visual generator. Randomised typography removes template matching but leaves orthography correlated with word identity, which cannot be solved within a reading paradigm.

Only a multimodal paradigm can ultimately answer the question of the visual confound, which is why we intend to follow up the present study with a passive-listening paradigm. If a decoder trained on reading transfers to listening, or if a jointly-trained decoder shows a shared component, then the shared component cannot be visual word form, because listening has none. Evidence for modality-general semantic codes \citep{correia2014brain, lin2022neural} provides cause for optimism, but it remains to be demonstrated at EEG signal-to-noise ratios.

\subsection{Limitations}
\label{sec:limitations}

\paragraph{Single-subject analysis}
All results presented here come from one densely-sampled individual. The cross-subject analysis using a 60-participant cohort is in preparation and will address both generalisation to unseen participants and whether pooling improves within-participant decoding, as it did in \citet{defossez2023decoding}.

\paragraph{Winner's curse}
Metrics presented in this report were computed on a validation set that also drove checkpoint selection and configuration selection. Absolute magnitudes are therefore upper bounds; cross-condition orderings are more reliable than the values themselves.

\paragraph{Task transfer}
Silent reading is a proxy for inner speech, and RSVP reading is not natural reading. Presenting words at a fixed location one at a time eliminates saccades and parafoveal preview, which is a considerable methodological advantage (precise timing, no eye-movement artefacts confounded with word identity), but simultaneously a departure from ecological reading. The paradigm should be understood as a compliant, scalable proxy, not as a naturalistic one.

\paragraph{Protocol heterogeneity}
Two changes occurred during data collection: the attention task and the ISI. Repeated words and behavioural responses are excluded at the trial level, so the one-back manipulation cannot contaminate the decoded trials, but the change in task instruction remains a source of variance. The ISI contrast is confounded with data volume and is not interpreted as an ISI effect.

\paragraph{Evaluation is retrieval, not generation}
Like all contrastive decoders \citep{defossez2023decoding, sato2024scaling}, our model identifies the most likely candidate from a supplied set rather than generating text. Given the current state of the EEG-to-text field (data limited, low signal to noise ratio) and challenges inherent to language modelling, contrastive learning is the practically preferable approach. In the future, methods already employed in the invasive literature, such as CTC-trained sequence models combined with a language model, might become useful in the non-invasive field as well \citep{card2024accurate, metzger2023high}.

\paragraph{Hardware limitations}
The 19-electrode configuration used in the present study is a low-density montage. Moreover, we used dry electrodes and a comparatively low sampling rate (600\,Hz). Higher-density wet-electrode EEG might reach higher accuracies. However, our hardware choice aligns with our belief in an experimental paradigm that prioritises scalability and subject compliance. We used an EEG device that is easier and quicker to set up and more comfortable to wear for extended sessions than a typical, wet-electrode research EEG device.

\subsection{Ongoing and future work}
\label{sec:future-work}

\paragraph{Cross-subject modelling}
A cross-subject analysis (60 participants, data collection ongoing) is in preparation. We plan to test whether multi-subject training improves decoding as it did for M/EEG speech perception \citep{defossez2023decoding}, and whether a model pretrained at scale (either on deep subject data or cross-subject pooled data) can be calibrated to a new participant (as \citealp{sato2024scaling} hypothesised).

\paragraph{Generalisability to inner monologue}
A multimodal dataset is required to test whether semantic language processing can be decoded independently of stimulus-induced sensory processing (see Section~\ref{sec:visual-confound}). We suggest that an experimental paradigm that prioritises scalability and subject compliance is preferable, which is why we intend to use a passive listening condition, where participants listen to audiobooks (using the same or similar text as in the present study), interrupted by comprehension questions to ensure subject compliance. We propose a multi-stage research agenda towards non-invasive brain-computer interfaces. First, modality-specific decoding needs to be established (using large silent reading and passive listening datasets). Second, cross-modal, non-sensory language decoding needs to be achieved (using combined reading and listening data). Finally, if the first two steps succeed, a smaller inner-monologue dataset (using a repetitive or generative inner-speech paradigm, as discussed by \citealp{csaky2025towards}) can be used to fine-tune a model for inner-monologue decoding. We suggest an analogy to the development of large language models (LLMs). Transformer-based LLMs required an unprecedentedly large dataset to train. The practical solution was to use almost the entire internet as a text corpus for pretraining. To make LLMs actually useful, expensive-to-obtain fine-tuning datasets were required (e.g.\ human preference data). Similarly, we propose to prioritise scalability and subject compliance for obtaining large proxy datasets, before fine-tuning on a smaller, difficult-to-obtain inner-monologue dataset. Reading and listening are obvious choices for proxy tasks, but watching movies or interactive video games involving language might also be considered (although the latter would struggle with motor activity confounds).

\subsection{Conclusion}
\label{sec:conclusion}

For developing non-invasive inner-monologue decoding, a large, labelled inner-monologue dataset would be ideal. However, such a dataset cannot practically be obtained. But there is a viable proxy. We have shown that using a modestly-sized dataset (${\sim}49$\,h) from a single participant, an open-vocabulary contrastive decoder extracts word-level information from EEG during naturalistic silent reading. Our result is not attributable to narrative topic tracking or to positional priors, is not confined to frequent words, and is still improving log-linearly with the amount of data collected. The present results are the first step in a multi-stage research agenda aimed towards inner-monologue decoding. For developing non-invasive language decoding, we suggest the use of proxy tasks that prioritise scalability and subject compliance.

\bibliographystyle{plainnat}
\bibliography{references}

\clearpage
\appendix
\setcounter{section}{-1}
\renewcommand{\thesection}{S\arabic{section}}
\setcounter{table}{0}
\renewcommand{\thetable}{S\arabic{table}}
\setcounter{figure}{0}
\renewcommand{\thefigure}{S\arabic{figure}}

\begin{center}
  {\LARGE\bfseries Supplementary Material}
\end{center}

\section{EEG hardware}
\label{sec:s0}

EEG was recorded from 19 scalp electrodes using a dry-electrode system (DSI-24 from Wearable Sensing), at a sampling rate of 600\,Hz. Electrode positions were: Fp1, Fp2, F3, F4, F7, F8, Fz, C3, C4, Cz, T3, T4, T5, T6, P3, P4, Pz, O1, O2 of the 10-20 layout. The DSI-24 system also measures from two earclip sensors placed on the earlobes (A1, A2).

\section{Metric glossary}
\label{sec:s1}

All metrics are word-grouped top-10 retrieval within fixed 512-trial candidate pools, averaged over five independent pool draws, and all ``gains'' are model score minus an empirical permutation baseline (10 permutations), so that chance is zero. Gains are quoted in percentage points (pp) throughout the report (Section~\ref{sec:metrics}).

\begin{table}[htb]
  \centering
  \caption{Metric glossary.}
  \label{tab:s-glossary}
  \footnotesize
  \setlength{\tabcolsep}{4pt}
  \begin{tabular}{p{3.2cm}p{3.6cm}p{3.4cm}p{4.2cm}}
    \toprule
    Paper term & Pool construction & Null / baseline & What it measures \\
    \midrule
    \textbf{Overall retrieval gain} & Random samples of 512 validation trials (mixes recording runs) & Permutation of predictions within the pool & Total decoding signal, including passage-level context \\
    \addlinespace
    \textbf{Within-run retrieval gain} \emph{(primary)} & Consecutive trials from a single recording run, chunked to $\le 512$ & Permutation within the same run (within-run EEG shuffle) & Discrimination among words in the same passage; excludes cross-run topic and run-identity drift \\
    \addlinespace
    \textbf{Context-tracking gain} & Random cross-run pools (as overall) & Each trial's EEG prediction replaced by a run-mate's & The passage-tracking component of the overall gain \\
    \addlinespace
    \textbf{Context-independent gain} & n/a & n/a & Overall gain $-$ context-tracking gain \\
    \addlinespace
    \textbf{Headroom-normalised accuracy} & as overall & as overall & (accuracy $-$ baseline) / (1 $-$ baseline) \\
    \addlinespace
    Rare / mid / frequent bin gains & as parent regime & per-bin empirical permutation rate & Frequency-resolved profile \\
    \bottomrule
  \end{tabular}
\end{table}

Position probes (Sections~\ref{sec:position-probes}, \ref{sec:results-position}): \textbf{position-only}, \textbf{preceding-EEG} and \textbf{current-trial} are the nested decomposition of the within-run gain and sum to it (unlike the pooling regimes above, which must not be summed); \textbf{position-corrected} $=$ within-run gain $-$ position-only. Only the position-only term is non-neural. The dispersion diagnostic validates the position-only null (Supplementary~\ref{sec:s6-caveats}).

Word grouping: before evaluation, each stimulus word is reduced to a ``core'' form by stripping leading and trailing punctuation and case-folding, so that ``Horse'', ``horse'' and ``horse,'' are treated as one candidate. Probability mass from all trials sharing a core word is summed before ranking.

Checkpoint selection uses the within-run gain, averaged over the five pool draws. Validation loss is logged but not used for selection; loss-based selection would systematically under-select the configurations of scientific interest. Loss is not comparable across the within-run negative-masking factor, which changes the number of terms in the softmax.

\section{Model architecture}
\label{sec:s2}

\paragraph{EEG encoder}
Input is a fixed-length window of $C \times T$ samples ($C = 19$ or 15 channels; $T = 300$, 420 or 540 samples at 600\,Hz for 0.5/0.7/0.9\,s windows).

\begin{itemize}
  \item \emph{Global pathway}: four masked 1-D convolutional blocks with output channels (48, 96, 192, 384), kernel sizes (11, 9, 7, 5) and strides (1, 2, 2, 2), each block being Conv1d $\rightarrow$ masked BatchNorm $\rightarrow$ ELU $\rightarrow$ 20\% dropout. Time is then collapsed by a learnable attention pooling (a linear query producing softmax weights over valid timepoints).
  \item \emph{Local pathway}: three stride-1 blocks with output channels (48, 96, 192) and kernel sizes (11, 9, 7), followed by masked adaptive average pooling into 6 temporal bins, preserving coarse within-window timing.
  \item \emph{Head}: concatenation of the two pathway outputs $\rightarrow$ two fully-connected layers of 512 units (BatchNorm, ELU, 20\% dropout) $\rightarrow$ linear projection to a 256-dimensional per-trial feature vector.
\end{itemize}

All normalisation and pooling operations are mask-aware, computed over valid timepoints only, so zero-padding cannot leak into a trial's feature vector.

\paragraph{Sequence model (optional)}
Four \texttt{torch.nn.TransformerEncoderLayer} blocks with d\_model $=$ 256, 8 attention heads, feed-forward dimension 1{,}024, dropout 0.4, GELU activation, pre-norm, learned positional embeddings (maximum sequence length 1{,}400), final LayerNorm. A causal mask restricts position $i$ to attend to positions $j \le i$, mirroring the causal attention of the language model that produced the targets.

\paragraph{Excluded trials}
Trials failing the amplitude criterion and target/false-alarm trials (in the early data subset using a one-back attention task instead of comprehension questions) are never passed through the encoder (so they cannot contaminate BatchNorm statistics) and their position in the sequence is filled with a learnable \texttt{[MASK]} token, trained end-to-end. The transformer therefore ``knows'' that a word occurred at that position but that its EEG is unavailable, and the alignment with the target sequence is preserved. These positions are excluded from the loss and from all evaluation.

\paragraph{Projections and loss}
A bias-free linear layer maps encoder/transformer output to a 256-dimensional shared space; a second bias-free linear layer maps the 4{,}096-dimensional LLM target into the same space. Both are L2-normalised. The symmetric CLIP cross-entropy \citep{radford2021learning} is computed over all valid positions in the batch, with a learnable log-temperature initialised at $\log(1/0.07)$ and clamped at $\exp \le 100$.

\paragraph{Batching}
One batch item is a complete recording run (${\sim}600$ trials), so a batch of 4 runs contributes ca.\ 2{,}200 valid in-batch negatives (assuming ca.\ 90\% valid trials).

\section{Hyperparameter grids}
\label{sec:s3}

\textbf{Fixed across all reported sweeps:} deep participant only; target embeddings from Llama-3.1-8B, 4{,}096-dimensional; projection dimension 256; 100 epochs; AdamW with weight decay 0.1; cosine schedule with 5\% warm-up; no gradient clipping; band-pass 1--80\,Hz (4th-order Butterworth); band-stop $\pm1$\,Hz around 50\,Hz mains (4th-order Butterworth); amplitude rejection at 100\,\muV{}; target/false-alarm trials excluded; validation fraction 0.2; content-aware text split with 10-gram overlap threshold; evaluation pool size 512; 10 permutations; 5 evaluation seeds; 3 frequency bins; position probes (single-pass all-masked, and grouped leave-one-out mask-current) evaluated for every fit.

\begin{table}[htb]
  \centering
  \caption{Sweep 1, 576 fits $= 2 \times 3 \times 3 \times 2 \times 2 \times 2 \times 2 \times 2$.}
  \label{tab:s-sweep1-grid}
  \small
  \begin{tabular}{ll}
    \toprule
    Factor & Levels \\
    \midrule
    LLM embedding layer & 0, 20 \\
    Window lock & onset, offset, centre \\
    ISI subset & \{0\,ms\}, \{100\,ms\}, \{0, 100\,ms\} \\
    Sequence model & none, transformer \\
    Normalisation & scale-only, baseline subtraction $+$ scale \\
    Temporal-shift augmentation & 0, $\pm100$\,ms \\
    Mask within-run negatives & False, True \\
    Trainable temperature & False, True \\
    \emph{(fixed)} window length & 0.5\,s \\
    \emph{(fixed)} batch size / learning rate & 4 / 1e-4 \\
    \bottomrule
  \end{tabular}
\end{table}

\textbf{Sweep 2 (data scaling), $2 \times 10$ fits.} Fixed configuration: onset lock, 0.5\,s window, no augmentation, scale-only normalisation, both ISIs, no negative masking, trainable temperature, batch 4, learning rate 1e-4. Swept: the fraction of training runs retained, $\in \{0.1, \ldots, 1.0\}$, separately for (layer 20 $+$ transformer) and (layer 0, no transformer). Only training runs are subsampled; the content-aware split is computed beforehand, so the same validation trials are scored at every ratio.

\begin{table}[htb]
  \centering
  \caption{Sweep 3, 192 fits $= 2 \times 3 \times 2 \times 2 \times 2 \times 2 \times 2$.}
  \label{tab:s-sweep3-grid}
  \small
  \begin{tabular}{ll}
    \toprule
    Factor & Levels \\
    \midrule
    LLM embedding layer & 0, 20 \\
    Window length & 0.5, 0.7, 0.9\,s \\
    Temporal-shift augmentation & 0, $\pm50$\,ms \\
    \textbf{Channel set} & all 19; O1/O2/T5/T6 removed \\
    Sequence model & none, transformer \\
    Batch size & 4, 8 \\
    Learning rate & 1e-4, 1e-5 \\
    \emph{(fixed)} window lock / ISI / normalisation & onset / both / scale-only \\
    \bottomrule
  \end{tabular}
\end{table}

\section{Absolute accuracies}
\label{sec:s4}

We report gains rather than raw accuracies because the raw top-10 accuracy in a 512-trial pool is dominated by the frequency structure of English prose: a frequent word occupies many candidate slots and accumulates probability mass irrespective of the EEG. The permutation baseline absorbs this, which is why chance is zero for every gain. The below raw accuracy numbers are provided for completeness, and are not directly comparable across studies without matching the pool construction.

$V$ $=$ mean number of \textbf{distinct core words} per evaluation pool, which is not the pool size: words are grouped before ranking, so $V$ is smaller than the pool size of 512. $10/V$ is the `uniform over candidate words' reference and understates chance, because a word filling many trial slots draws mass from all of them; the measured permutation baseline (the `implied empirical chance' column) sits above it and is the figure to quote. The gain column is in percentage points, so gain $=$ implied model top-10 $-$ implied empirical chance (up to rounding); norm-acc is the headroom-normalised accuracy, a ratio rather than a percentage.

\begin{table}[htb]
  \centering
  \caption{Sweep 1, random cross-run pools: absolute accuracies. Across all 576 fits: mean $V = 299$ distinct core words per pool, so the naive uniform-over-words reference is $10/V = 3.34\%$, against a measured empirical chance of 14.31\% and a model top-10 of 25.96\%.}
  \label{tab:s4-sweep1-overall}
  \footnotesize
  \setlength{\tabcolsep}{4pt}
  \begin{tabular}{lcccccccc}
    \toprule
    Configuration & \makecell[c]{$n$\\(gain)} & \makecell[c]{$n$\\(abs)} & \makecell[c]{Gain\\(pp)} & norm-acc & \makecell[c]{Implied\\empirical\\chance} & \makecell[c]{Implied\\model\\top-10} & \makecell[c]{$V$ (distinct\\words / pool)} & \makecell[c]{$10/V$\\reference} \\
    \midrule
    all fits & 576 & 576 & 11.64 & 0.1352 & 14.31\% & 25.96\% & 299 & 3.34\% \\
    noTF\_L0 & 144 & 144 & 7.92 & 0.0930 & 15.09\% & 23.01\% & 299 & 3.34\% \\
    noTF\_L20 & 144 & 144 & 10.95 & 0.1272 & 14.19\% & 25.14\% & 300 & 3.34\% \\
    TF\_L0 & 144 & 144 & 7.90 & 0.0921 & 14.42\% & 22.32\% & 299 & 3.34\% \\
    TF\_L20 & 144 & 144 & 19.81 & 0.2286 & 13.54\% & 33.35\% & 300 & 3.34\% \\
    \bottomrule
  \end{tabular}
\end{table}

\begin{table}[htb]
  \centering
  \caption{Sweep 1, within-run / strict (by-run pools): absolute accuracies.}
  \label{tab:s4-sweep1-withinrun}
  \footnotesize
  \setlength{\tabcolsep}{4pt}
  \begin{tabular}{lcccccccc}
    \toprule
    Configuration & \makecell[c]{$n$\\(gain)} & \makecell[c]{$n$\\(abs)} & \makecell[c]{Gain\\(pp)} & norm-acc & \makecell[c]{Implied\\empirical\\chance} & \makecell[c]{Implied\\model\\top-10} & \makecell[c]{$V$ (distinct\\words / pool)} & \makecell[c]{$10/V$\\reference} \\
    \midrule
    all fits & 576 & 576 & 6.74 & 0.0822 & 18.30\% & 25.03\% & 246 & 4.07\% \\
    noTF\_L0 & 144 & 144 & 7.48 & 0.0904 & 17.43\% & 24.91\% & 246 & 4.07\% \\
    noTF\_L20 & 144 & 144 & 5.47 & 0.0677 & 19.69\% & 25.16\% & 246 & 4.07\% \\
    TF\_L0 & 144 & 144 & 6.18 & 0.0740 & 16.53\% & 22.71\% & 246 & 4.07\% \\
    TF\_L20 & 144 & 144 & 7.81 & 0.0965 & 19.54\% & 27.35\% & 246 & 4.07\% \\
    \bottomrule
  \end{tabular}
\end{table}

\begin{table}[htb]
  \centering
  \caption{Sweep 3, random cross-run pools: absolute accuracies. Across all 192 fits: mean $V = 301$ distinct core words per pool, so the naive uniform-over-words reference is $10/V = 3.32\%$, against a measured empirical chance of 15.75\% and a model top-10 of 29.28\%.}
  \label{tab:s4-sweep3-overall}
  \footnotesize
  \setlength{\tabcolsep}{4pt}
  \begin{tabular}{lcccccccc}
    \toprule
    Configuration & \makecell[c]{$n$\\(gain)} & \makecell[c]{$n$\\(abs)} & \makecell[c]{Gain\\(pp)} & norm-acc & \makecell[c]{Implied\\empirical\\chance} & \makecell[c]{Implied\\model\\top-10} & \makecell[c]{$V$ (distinct\\words / pool)} & \makecell[c]{$10/V$\\reference} \\
    \midrule
    all fits & 192 & 192 & 13.54 & 0.1583 & 15.75\% & 29.28\% & 301 & 3.32\% \\
    noTF\_L0 & 48 & 48 & 7.11 & 0.0857 & 17.78\% & 24.89\% & 301 & 3.32\% \\
    noTF\_L20 & 48 & 48 & 13.69 & 0.1604 & 15.32\% & 29.01\% & 302 & 3.31\% \\
    TF\_L0 & 48 & 48 & 7.93 & 0.0945 & 16.46\% & 24.38\% & 301 & 3.32\% \\
    TF\_L20 & 48 & 48 & 25.42 & 0.2925 & 13.44\% & 38.86\% & 301 & 3.32\% \\
    all\_channels & 96 & 96 & 15.24 & 0.1776 & 15.21\% & 30.45\% & 301 & 3.32\% \\
    no\_visual\_channels & 96 & 96 & 11.83 & 0.1390 & 16.29\% & 28.12\% & 302 & 3.32\% \\
    \bottomrule
  \end{tabular}
\end{table}

\begin{table}[htb]
  \centering
  \caption{Sweep 3, within-run / strict (by-run pools): absolute accuracies.}
  \label{tab:s4-sweep3-withinrun}
  \footnotesize
  \setlength{\tabcolsep}{4pt}
  \begin{tabular}{lcccccccc}
    \toprule
    Configuration & \makecell[c]{$n$\\(gain)} & \makecell[c]{$n$\\(abs)} & \makecell[c]{Gain\\(pp)} & norm-acc & \makecell[c]{Implied\\empirical\\chance} & \makecell[c]{Implied\\model\\top-10} & \makecell[c]{$V$ (distinct\\words / pool)} & \makecell[c]{$10/V$\\reference} \\
    \midrule
    all fits & 192 & 192 & 7.76 & 0.0963 & 19.96\% & 27.72\% & 244 & 4.09\% \\
    noTF\_L0 & 48 & 48 & 7.37 & 0.0919 & 20.34\% & 27.71\% & 244 & 4.09\% \\
    noTF\_L20 & 48 & 48 & 6.71 & 0.0839 & 20.88\% & 27.59\% & 245 & 4.09\% \\
    TF\_L0 & 48 & 48 & 6.93 & 0.0849 & 18.73\% & 25.66\% & 244 & 4.09\% \\
    TF\_L20 & 48 & 48 & 10.04 & 0.1244 & 19.89\% & 29.93\% & 244 & 4.09\% \\
    all\_channels & 96 & 96 & 9.24 & 0.1140 & 19.39\% & 28.63\% & 244 & 4.09\% \\
    no\_visual\_channels & 96 & 96 & 6.28 & 0.0785 & 20.53\% & 26.82\% & 244 & 4.09\% \\
    \bottomrule
  \end{tabular}
\end{table}

\section{Complete result tables}
\label{sec:s5}

\begin{table}[htb]
  \centering
  \caption{Sweep 1: decomposition by factor (mean $\pm$ SD). $n = 288$ per level for two-level factors, 192 for three-level factors. Gains in percentage points (pp); norm-acc is a ratio.}
  \label{tab:s5-sweep1-factors}
  \footnotesize
  \setlength{\tabcolsep}{3.5pt}
  \begin{tabular}{llcccccc}
    \toprule
    Factor & Level & \makecell[c]{Within-run\\gain (pp)} & \makecell[c]{Overall\\gain (pp)} & \makecell[c]{Context-\\tracking\\gain (pp)} & \makecell[c]{Context-\\independent\\gain (pp)} & norm-acc & val-loss \\
    \midrule
    Embedding layer & 0 & $6.83 \pm 2.22$ & $7.91 \pm 2.08$ & $1.46 \pm 0.81$ & $6.45 \pm 2.07$ & 0.0925 & 7.645 \\
    & 20 & $6.64 \pm 3.13$ & $15.38 \pm 6.57$ & $5.24 \pm 1.71$ & $10.14 \pm 5.80$ & 0.1779 & 6.648 \\
    Window lock & onset & $6.80 \pm 2.86$ & $11.74 \pm 6.34$ & $3.38 \pm 2.40$ & $8.37 \pm 4.88$ & 0.1363 & 7.138 \\
    & offset & $6.50 \pm 2.48$ & $11.42 \pm 6.00$ & $3.37 \pm 2.31$ & $8.05 \pm 4.57$ & 0.1327 & 7.169 \\
    & centre & $6.91 \pm 2.78$ & $11.77 \pm 6.10$ & $3.30 \pm 2.25$ & $8.47 \pm 4.74$ & 0.1367 & 7.133 \\
    ISI subset & 0\,ms & $6.86 \pm 2.67$ & $11.67 \pm 5.55$ & $3.37 \pm 2.18$ & $8.30 \pm 4.38$ & 0.1356 & 7.162 \\
    & 100\,ms & $6.14 \pm 2.51$ & $9.33 \pm 4.55$ & $2.38 \pm 1.47$ & $6.96 \pm 3.68$ & 0.1093 & 7.364 \\
    & both & $7.21 \pm 2.84$ & $13.93 \pm 7.16$ & $4.30 \pm 2.72$ & $9.63 \pm 5.56$ & 0.1609 & 6.914 \\
    Sequence model & none & $6.48 \pm 2.46$ & $9.43 \pm 3.00$ & $2.82 \pm 2.17$ & $6.62 \pm 2.49$ & 0.1101 & 7.397 \\
    & transformer & $7.00 \pm 2.93$ & $13.86 \pm 7.53$ & $3.88 \pm 2.35$ & $9.98 \pm 5.73$ & 0.1603 & 6.896 \\
    Normalisation & baseline $+$ scale & $6.23 \pm 2.63$ & $11.07 \pm 6.04$ & $3.33 \pm 2.31$ & $7.74 \pm 4.62$ & 0.1287 & 7.187 \\
    & scale only & $7.24 \pm 2.71$ & $12.21 \pm 6.20$ & $3.36 \pm 2.33$ & $8.85 \pm 4.78$ & 0.1417 & 7.106 \\
    Temporal shift & 0 & $8.38 \pm 2.36$ & $13.07 \pm 5.96$ & $3.16 \pm 2.25$ & $9.91 \pm 4.53$ & 0.1512 & 7.131 \\
    & $\pm100$\,ms & $5.09 \pm 1.93$ & $10.22 \pm 5.99$ & $3.54 \pm 2.37$ & $6.69 \pm 4.36$ & 0.1193 & 7.162 \\
    Mask within-run neg. & False & $8.02 \pm 2.65$ & $12.91 \pm 7.49$ & $2.93 \pm 2.23$ & $9.97 \pm 5.61$ & 0.1501 & 7.502 \\
    & True & $5.45 \pm 2.10$ & $10.38 \pm 4.03$ & $3.76 \pm 2.33$ & $6.62 \pm 2.77$ & 0.1204 & 6.791 \\
    Trainable temperature & False & $6.76 \pm 2.71$ & $11.77 \pm 6.28$ & $3.39 \pm 2.32$ & $8.38 \pm 4.82$ & 0.1371 & 7.097 \\
    & True & $6.71 \pm 2.71$ & $11.52 \pm 6.01$ & $3.30 \pm 2.32$ & $8.22 \pm 4.63$ & 0.1334 & 7.196 \\
    Context condition & L0, no TF & $7.48 \pm 2.02$ & $7.92 \pm 1.86$ & $1.02 \pm 0.48$ & $6.89 \pm 1.86$ & 0.0930 & 7.652 \\
    & L0, TF & $6.18 \pm 2.23$ & $7.90 \pm 2.28$ & $1.89 \pm 0.85$ & $6.01 \pm 2.18$ & 0.0921 & 7.638 \\
    & L20, no TF & $5.47 \pm 2.45$ & $10.95 \pm 3.15$ & $4.61 \pm 1.65$ & $6.34 \pm 2.98$ & 0.1272 & 7.142 \\
    & L20, TF & $7.81 \pm 3.30$ & $19.81 \pm 6.10$ & $5.86 \pm 1.54$ & $13.94 \pm 5.43$ & 0.2286 & 6.154 \\
    \bottomrule
  \end{tabular}
\end{table}

\begin{table}[htb]
  \centering
  \caption{Sweep 1: within-run gain by frequency bin. Gains in percentage points (pp).}
  \label{tab:s5-sweep1-bins}
  \small
  \begin{tabular}{llccc}
    \toprule
    Factor & Level & Rare & Mid & Frequent \\
    \midrule
    Embedding layer & 0 & $2.81 \pm 1.53$ & $2.80 \pm 1.44$ & $7.43 \pm 2.39$ \\
    & 20 & $4.75 \pm 3.10$ & $4.81 \pm 2.96$ & $6.92 \pm 3.16$ \\
    Window lock & onset & $3.27 \pm 2.58$ & $3.41 \pm 2.51$ & $7.31 \pm 3.00$ \\
    & offset & $3.98 \pm 2.52$ & $3.92 \pm 2.37$ & $6.88 \pm 2.56$ \\
    & centre & $4.08 \pm 2.72$ & $4.07 \pm 2.68$ & $7.34 \pm 2.85$ \\
    ISI subset & 0\,ms & $4.30 \pm 2.56$ & $4.29 \pm 2.53$ & $7.26 \pm 2.79$ \\
    & 100\,ms & $2.75 \pm 2.07$ & $2.83 \pm 2.03$ & $6.67 \pm 2.67$ \\
    & both & $4.29 \pm 2.89$ & $4.28 \pm 2.72$ & $7.59 \pm 2.90$ \\
    Sequence model & none & $3.36 \pm 1.77$ & $3.41 \pm 1.78$ & $6.94 \pm 2.64$ \\
    & transformer & $4.19 \pm 3.22$ & $4.19 \pm 3.06$ & $7.41 \pm 2.96$ \\
    Normalisation & baseline $+$ scale & $3.47 \pm 2.44$ & $3.50 \pm 2.42$ & $6.64 \pm 2.73$ \\
    & scale only & $4.09 \pm 2.78$ & $4.10 \pm 2.62$ & $7.71 \pm 2.79$ \\
    Temporal shift & 0 & $4.76 \pm 2.81$ & $4.74 \pm 2.65$ & $8.92 \pm 2.42$ \\
    & $\pm100$\,ms & $2.80 \pm 2.00$ & $2.87 \pm 2.02$ & $5.43 \pm 1.96$ \\
    Context condition & L0, no TF & $3.22 \pm 1.49$ & $3.17 \pm 1.42$ & $8.12 \pm 2.18$ \\
    & L0, TF & $2.40 \pm 1.46$ & $2.42 \pm 1.36$ & $6.74 \pm 2.40$ \\
    & L20, no TF & $3.51 \pm 2.00$ & $3.64 \pm 2.05$ & $5.75 \pm 2.53$ \\
    & L20, TF & $5.98 \pm 3.50$ & $5.97 \pm 3.26$ & $8.08 \pm 3.30$ \\
    \bottomrule
  \end{tabular}
\end{table}

\paragraph{Sweep 2 --- data scaling}
Per-ratio selected-epoch scalars for the two scaling arms. The log-linear fits over the ten ratios, including the position-probe traces, are reported in Table~\ref{tab:scaling}. One fit per ratio: the slope is the reportable quantity, and the per-ratio levels are selection-optimistic upper bounds. In fits trained on small fractions of the data, the validation loss reached its minimum early and rose thereafter (a sign of over-fitting), while the within-run gain continued to improve; gain-based checkpoint selection (Section~\ref{sec:model}) therefore matters most at small data volumes.

\begin{table}[htb]
  \centering
  \caption{Sweep 3: decomposition by factor (mean $\pm$ SD). Gains in percentage points (pp); norm-acc is a ratio. Distribution across all 192 fits: within-run gain $7.76 \pm 2.89$\,pp (median 7.52, min 1.99, max 16.53); overall gain $13.54 \pm 8.52$\,pp; context-tracking gain $3.41 \pm 2.64$\,pp.}
  \label{tab:s5-sweep3-factors}
  \footnotesize
  \setlength{\tabcolsep}{3.5pt}
  \begin{tabular}{llcccccc}
    \toprule
    Factor & Level & \makecell[c]{Within-run\\gain (pp)} & \makecell[c]{Overall\\gain (pp)} & \makecell[c]{Context-\\tracking\\gain (pp)} & \makecell[c]{Context-\\independent\\gain (pp)} & norm-acc & val-loss \\
    \midrule
    Embedding layer & 0 & $7.15 \pm 2.17$ & $7.52 \pm 2.35$ & $1.06 \pm 0.48$ & $6.46 \pm 2.19$ & 0.0901 & 7.969 \\
    & 20 & $8.38 \pm 3.36$ & $19.55 \pm 8.20$ & $5.76 \pm 1.60$ & $13.79 \pm 6.81$ & 0.2264 & 7.134 \\
    Window length & 0.5\,s & $6.86 \pm 2.73$ & $12.24 \pm 8.06$ & $3.26 \pm 2.52$ & $8.97 \pm 5.89$ & 0.1434 & 7.612 \\
    & 0.7\,s & $8.32 \pm 2.85$ & $14.14 \pm 8.78$ & $3.41 \pm 2.68$ & $10.74 \pm 6.44$ & 0.1651 & 7.542 \\
    & 0.9\,s & $8.11 \pm 2.90$ & $14.22 \pm 8.69$ & $3.57 \pm 2.73$ & $10.66 \pm 6.31$ & 0.1664 & 7.499 \\
    Temporal shift & 0 & $8.31 \pm 2.88$ & $13.93 \pm 8.41$ & $3.30 \pm 2.56$ & $10.63 \pm 6.19$ & 0.1625 & 7.565 \\
    & $\pm50$\,ms & $7.22 \pm 2.80$ & $13.14 \pm 8.66$ & $3.52 \pm 2.72$ & $9.61 \pm 6.28$ & 0.1540 & 7.537 \\
    \textbf{Channel set} & all 19 & $9.24 \pm 2.66$ & $15.24 \pm 8.63$ & $3.48 \pm 2.68$ & $11.75 \pm 6.29$ & 0.1776 & 7.480 \\
    & O1/O2/T5/T6 removed & $6.28 \pm 2.29$ & $11.83 \pm 8.09$ & $3.34 \pm 2.61$ & $8.49 \pm 5.78$ & 0.1390 & 7.622 \\
    Sequence model & none & $7.04 \pm 2.49$ & $10.40 \pm 4.91$ & $2.74 \pm 2.29$ & $7.66 \pm 3.21$ & 0.1230 & 7.741 \\
    & transformer & $8.49 \pm 3.08$ & $16.67 \pm 10.10$ & $4.08 \pm 2.80$ & $12.59 \pm 7.46$ & 0.1935 & 7.362 \\
    Batch size & 4 & $8.21 \pm 2.88$ & $13.66 \pm 8.26$ & $3.14 \pm 2.33$ & $10.51 \pm 6.23$ & 0.1596 & 7.312 \\
    & 8 & $7.31 \pm 2.84$ & $13.41 \pm 8.81$ & $3.68 \pm 2.90$ & $9.73 \pm 6.26$ & 0.1569 & 7.790 \\
    Learning rate & 1e-4 & $9.39 \pm 2.81$ & $16.66 \pm 9.66$ & $4.04 \pm 2.99$ & $12.62 \pm 7.04$ & 0.1918 & 7.600 \\
    & 1e-5 & $6.14 \pm 1.89$ & $10.41 \pm 5.73$ & $2.79 \pm 2.06$ & $7.62 \pm 4.02$ & 0.1248 & 7.502 \\
    Context condition & L0, no TF & $7.37 \pm 2.33$ & $7.11 \pm 2.57$ & $0.65 \pm 0.29$ & $6.46 \pm 2.38$ & 0.0857 & 7.952 \\
    & L0, TF & $6.93 \pm 1.99$ & $7.93 \pm 2.05$ & $1.48 \pm 0.19$ & $6.45 \pm 2.00$ & 0.0945 & 7.986 \\
    & L20, no TF & $6.71 \pm 2.62$ & $13.69 \pm 4.46$ & $4.84 \pm 1.22$ & $8.85 \pm 3.51$ & 0.1604 & 7.529 \\
    & L20, TF & $10.04 \pm 3.21$ & $25.42 \pm 6.76$ & $6.68 \pm 1.40$ & $18.73 \pm 5.61$ & 0.2925 & 6.738 \\
    \bottomrule
  \end{tabular}
\end{table}

\begin{table}[htb]
  \centering
  \caption{Sweep 3: within-run gain by frequency bin. Gains in percentage points (pp).}
  \label{tab:s5-sweep3-bins}
  \small
  \begin{tabular}{llccc}
    \toprule
    Factor & Level & Rare & Mid & Frequent \\
    \midrule
    Embedding layer & 0 & $3.39 \pm 1.26$ & $3.51 \pm 1.27$ & $7.63 \pm 2.32$ \\
    & 20 & $7.05 \pm 4.17$ & $6.85 \pm 3.81$ & $8.56 \pm 3.30$ \\
    Window length & 0.5\,s & $3.64 \pm 2.75$ & $3.80 \pm 2.71$ & $7.26 \pm 2.78$ \\
    & 0.7\,s & $5.90 \pm 3.56$ & $5.70 \pm 3.28$ & $8.65 \pm 2.85$ \\
    & 0.9\,s & $6.11 \pm 3.84$ & $6.04 \pm 3.41$ & $8.38 \pm 2.87$ \\
    Temporal shift & 0 & $5.62 \pm 3.74$ & $5.61 \pm 3.41$ & $8.65 \pm 2.87$ \\
    & $\pm50$\,ms & $4.81 \pm 3.38$ & $4.75 \pm 3.11$ & $7.54 \pm 2.80$ \\
    Channel set & all 19 & $6.11 \pm 3.96$ & $6.11 \pm 3.49$ & $9.65 \pm 2.60$ \\
    & O1/O2/T5/T6 removed & $4.33 \pm 2.90$ & $4.25 \pm 2.79$ & $6.54 \pm 2.25$ \\
    Sequence model & none & $4.43 \pm 2.31$ & $4.49 \pm 2.18$ & $7.37 \pm 2.59$ \\
    & transformer & $6.00 \pm 4.37$ & $5.87 \pm 4.00$ & $8.82 \pm 2.99$ \\
    Context condition & L0, no TF & $3.63 \pm 1.33$ & $3.82 \pm 1.38$ & $7.84 \pm 2.49$ \\
    & L0, TF & $3.15 \pm 1.14$ & $3.21 \pm 1.08$ & $7.41 \pm 2.13$ \\
    & L20, no TF & $5.24 \pm 2.78$ & $5.17 \pm 2.60$ & $6.91 \pm 2.64$ \\
    & L20, TF & $8.86 \pm 4.54$ & $8.52 \pm 4.10$ & $10.22 \pm 3.08$ \\
    \bottomrule
  \end{tabular}
\end{table}

\begin{table}[htb]
  \centering
  \caption{Sweep 3: channel-ablation interactions (within-run gain). Gains in percentage points (pp). The ordering of every crossed factor is preserved across channel sets; the ablation is additive rather than interactive in sign, though its relative magnitude is larger in the non-contextual and no-transformer cells.}
  \label{tab:s5-sweep3-ablation}
  \small
  \begin{tabular}{llcc}
    \toprule
    Crossed factor & Level & All 19 channels & O1/O2/T5/T6 removed \\
    \midrule
    Embedding layer & 0 & $8.76 \pm 1.61$ & $5.54 \pm 1.26$ \\
    & 20 & $9.72 \pm 3.36$ & $7.03 \pm 2.81$ \\
    Sequence model & none & $8.58 \pm 2.22$ & $5.51 \pm 1.67$ \\
    & transformer & $9.91 \pm 2.91$ & $7.06 \pm 2.57$ \\
    Window length & 0.5\,s & $8.36 \pm 2.49$ & $5.35 \pm 2.08$ \\
    & 0.7\,s & $9.77 \pm 2.64$ & $6.87 \pm 2.28$ \\
    & 0.9\,s & $9.59 \pm 2.71$ & $6.63 \pm 2.29$ \\
    Batch size & 4 & $9.72 \pm 2.62$ & $6.71 \pm 2.29$ \\
    & 8 & $8.77 \pm 2.64$ & $5.86 \pm 2.24$ \\
    Learning rate & 1e-4 & $11.12 \pm 2.22$ & $7.65 \pm 2.19$ \\
    & 1e-5 & $7.36 \pm 1.47$ & $4.92 \pm 1.42$ \\
    \bottomrule
  \end{tabular}
\end{table}

\section{Position probes --- implementation caveats and per-factor tables}
\label{sec:s6}

All conventions as in Sections~\ref{sec:position-probes} and \ref{sec:results-position}: gains in percentage points (pp); the three probe terms are nested and sum to the within-run gain; position-corrected $=$ within-run $-$ position-only; a near-zero position-only gain is evidence only where dispersion is clearly above zero (\ref{sec:s6-caveats}; dispersion is a cosine distance, not a percentage); transformer-free cells are zero by construction and serve as a built-in correctness check on the probe machinery.

\subsection{Implementation caveats and validity of the position-only null}
\label{sec:s6-caveats}

The preceding-EEG level of the probe ladder (Section~\ref{sec:position-probes}, level ii) requires each trial's prediction to be calculated while that trial's own EEG is masked and all other trials' EEG is intact. Computing this exactly would require one transformer pass per position (${\sim}600$ per run). We approximate it with a grouped leave-one-out: positions are partitioned into 16 interleaved groups (position index modulo 16), one pass is run per group with the whole group masked, and each position's prediction is read from the pass in which it was itself masked. The property required for the probe to be valid (that the read-out position never sees its own EEG) holds. The approximation is that the same pass also masks the preceding positions congruent to the read-out position modulo 16, so ${\sim}1/16$ ($\approx 6\%$) of the preceding context is masked that the unablated model would have seen. The position-only probe masks every position in a single pass and involves no approximation.

Two caveats apply to the reading of the probe terms. First, because the grouped leave-one-out also masks a small fraction of the preceding positions, the preceding-EEG term is biased down and the current-trial term up by the same amount; the position-only and position-corrected terms are exact and unaffected, which is why the load-bearing claims rest on those two. Second, masking every trial is out of distribution, so a near-zero position-only gain is only evidence if the all-masked model still produces position-dependent output rather than collapsing to a constant (a constant prediction would score zero against a within-run shuffle, regardless of the model's decoding capability). We therefore report a dispersion diagnostic, the mean cosine distance of the position-only predictions from their own centroid, alongside the position-only gain. A near-zero gain is informative only where dispersion is above zero. For transformer-free configurations, both probes return exactly zero by construction (every position is processed independently, so both probes yield the same constant vector at every position); this serves as a built-in correctness check on the probe machinery rather than as an empirical result.

\begin{table}[htb]
  \centering
  \caption{Sweep 1: probes by architecture (mean over fits).}
  \label{tab:s6-sweep1-arch}
  \footnotesize
  \setlength{\tabcolsep}{4pt}
  \begin{tabular}{lccccccc}
    \toprule
    Level & $n$ & \makecell[c]{Within-run\\gain (pp)} & \makecell[c]{Position-\\only (pp)} & \makecell[c]{Preceding-\\EEG (pp)} & \makecell[c]{Current-\\trial (pp)} & \makecell[c]{Position-\\corrected (pp)} & Dispersion \\
    \midrule
    all fits & 576 & 6.74 & 0.70 & 0.49 & 5.55 & 6.04 & 0.0645 \\
    no transformer & 288 & 6.48 & 0.00 & 0.00 & 6.48 & 6.48 & 0.0000 \\
    transformer & 288 & 7.00 & 1.40 & 0.98 & 4.62 & 5.60 & 0.1290 \\
    \bottomrule
  \end{tabular}
\end{table}

\begin{table}[htb]
  \centering
  \caption{Sweep 1: position-corrected within-run gain per word-frequency bin.}
  \label{tab:s6-sweep1-bins}
  \small
  \begin{tabular}{lcccc}
    \toprule
    Level & $n$ & Rare & Mid & Frequent \\
    \midrule
    all fits & 576 & 3.20 & 3.24 & 6.46 \\
    L0, no TF & 144 & 3.22 & 3.17 & 8.12 \\
    L20, no TF & 144 & 3.51 & 3.64 & 5.75 \\
    L0, TF & 144 & 2.40 & 2.42 & 6.69 \\
    L20, TF & 144 & 3.69 & 3.72 & 5.26 \\
    \bottomrule
  \end{tabular}
\end{table}

\begin{table}[htb]
  \centering
  \caption{Sweep 3: probes by target type and architecture.}
  \label{tab:s6-sweep3-arch}
  \footnotesize
  \setlength{\tabcolsep}{4pt}
  \begin{tabular}{lccccccc}
    \toprule
    Level & $n$ & \makecell[c]{Within-run\\gain (pp)} & \makecell[c]{Position-\\only (pp)} & \makecell[c]{Preceding-\\EEG (pp)} & \makecell[c]{Current-\\trial (pp)} & \makecell[c]{Position-\\corrected (pp)} & Dispersion \\
    \midrule
    all fits & 192 & 7.76 & 0.73 & 0.58 & 6.46 & 7.03 & 0.0642 \\
    L0, no TF & 48 & 7.37 & 0.00 & 0.00 & 7.37 & 7.37 & 0.0000 \\
    L20, no TF & 48 & 6.71 & 0.00 & 0.00 & 6.71 & 6.71 & 0.0000 \\
    L0, TF & 48 & 6.93 & 0.07 & $-0.02$ & 6.88 & 6.86 & 0.1495 \\
    L20, TF & 48 & 10.04 & 2.85 & 2.34 & 4.86 & 7.20 & 0.1071 \\
    \bottomrule
  \end{tabular}
\end{table}

\begin{table}[htb]
  \centering
  \caption{Sweep 3: probes by channel set. The position-only term is essentially unchanged by the ablation while the current-trial term drops by ${\sim}37\%$: the ablation removes neural signal, not positional information, which is why the position-corrected gain falls in step with the raw within-run gain (Table~\ref{tab:ablation}).}
  \label{tab:s6-sweep3-channels}
  \footnotesize
  \setlength{\tabcolsep}{4pt}
  \begin{tabular}{lccccccc}
    \toprule
    Level & $n$ & \makecell[c]{Within-run\\gain (pp)} & \makecell[c]{Position-\\only (pp)} & \makecell[c]{Preceding-\\EEG (pp)} & \makecell[c]{Current-\\trial (pp)} & \makecell[c]{Position-\\corrected (pp)} & Dispersion \\
    \midrule
    all 19 channels & 96 & 9.24 & 0.77 & 0.56 & 7.92 & 8.48 & 0.0659 \\
    O1/O2/T5/T6 removed & 96 & 6.28 & 0.69 & 0.60 & 4.99 & 5.59 & 0.0624 \\
    \bottomrule
  \end{tabular}
\end{table}

\begin{table}[htb]
  \centering
  \caption{Sweep 3: position-corrected within-run gain per word-frequency bin.}
  \label{tab:s6-sweep3-bins}
  \small
  \begin{tabular}{lcccc}
    \toprule
    Level & $n$ & Rare & Mid & Frequent \\
    \midrule
    all fits & 192 & 4.67 & 4.66 & 7.34 \\
    L0, no TF & 48 & 3.63 & 3.82 & 7.84 \\
    L20, no TF & 48 & 5.24 & 5.17 & 6.91 \\
    L0, TF & 48 & 3.19 & 3.20 & 7.33 \\
    L20, TF & 48 & 6.62 & 6.46 & 7.28 \\
    all 19 channels & 96 & 5.53 & 5.58 & 8.85 \\
    O1/O2/T5/T6 removed & 96 & 3.80 & 3.74 & 5.83 \\
    \bottomrule
  \end{tabular}
\end{table}

\section{Control checks on technical factors}
\label{sec:s7}

\paragraph{Within-run negative masking (Sweep 1)}
Masking reduced the within-run gain in every crossed cell, and the ordering of the scientific factors was preserved in both masking conditions with one exception: For the non-contextual target, the overall gain was slightly \emph{higher} under masking (8.37 vs 7.45\,pp), reversing the direction seen for contextual targets (12.39 vs 18.37\,pp). The primary metric (within-run gain) did not reverse anywhere. Because masking changes the number of terms in the contrastive softmax, validation losses are not comparable across this factor.

Within-run gain in percentage points (pp).

\begin{table}[htb]
  \centering
  \caption{Sweep 1: within-run gain by within-run negative masking.}
  \label{tab:s7-masking}
  \small
  \begin{tabular}{llcc}
    \toprule
    Crossed factor & Level & Masking off & Masking on \\
    \midrule
    Embedding layer & 0 & $7.17 \pm 2.21$ & $6.49 \pm 2.18$ \\
    & 20 & $8.86 \pm 2.79$ & $4.42 \pm 1.39$ \\
    Sequence model & none & $7.41 \pm 2.02$ & $5.54 \pm 2.51$ \\
    & transformer & $8.62 \pm 3.05$ & $5.37 \pm 1.59$ \\
    Temporal shift & 0 & $9.88 \pm 1.93$ & $6.88 \pm 1.72$ \\
    & $\pm100$\,ms & $6.15 \pm 1.84$ & $4.03 \pm 1.34$ \\
    Context condition & L0, no TF & $7.70 \pm 2.03$ & $7.25 \pm 1.99$ \\
    & L20, TF & $10.61 \pm 2.37$ & $5.01 \pm 0.65$ \\
    \bottomrule
  \end{tabular}
\end{table}

\paragraph{Trainable temperature (Sweep 1)}
No effect on any metric in any cell; all differences were at or below ${\sim}0.1$\,pp on the within-run gain.

\paragraph{Temperature runaway}
Final values of the learnable temperature (logit scale): Sweep 1 mean 15.4 (max 19.3); Sweep 3 mean 15.7 (max 19.3), against a clamp of 100. No fit in any sweep approached runaway, so the temperature-dependent word-grouped metrics are undistorted throughout.

\paragraph{Training-budget diagnostics (Sweep 3)}
Selected checkpoint epoch was $80 \pm 15$ at learning rate 1e-4 and $91 \pm 7$ at 1e-5, and $83 \pm 13$ at batch 4 versus $87 \pm 13$ at batch 8, out of 100 epochs. Both the learning-rate and the batch-size effect are therefore partly attributable to an incomplete training budget rather than to an intrinsic optimum.

\paragraph{Training dynamics}
Validation loss frequently showed an extremum-then-reversal pattern in the non-contextual configurations (minimum around epoch 7--13, rising thereafter) while the within-run gain continued to improve, which is why checkpoint selection is not performed on loss.

\begin{figure}[tp]
  \centering
  \includegraphics[alt={Training curves of within-run gain and validation loss across epochs for Sweeps 1 and 3},width=\textwidth]{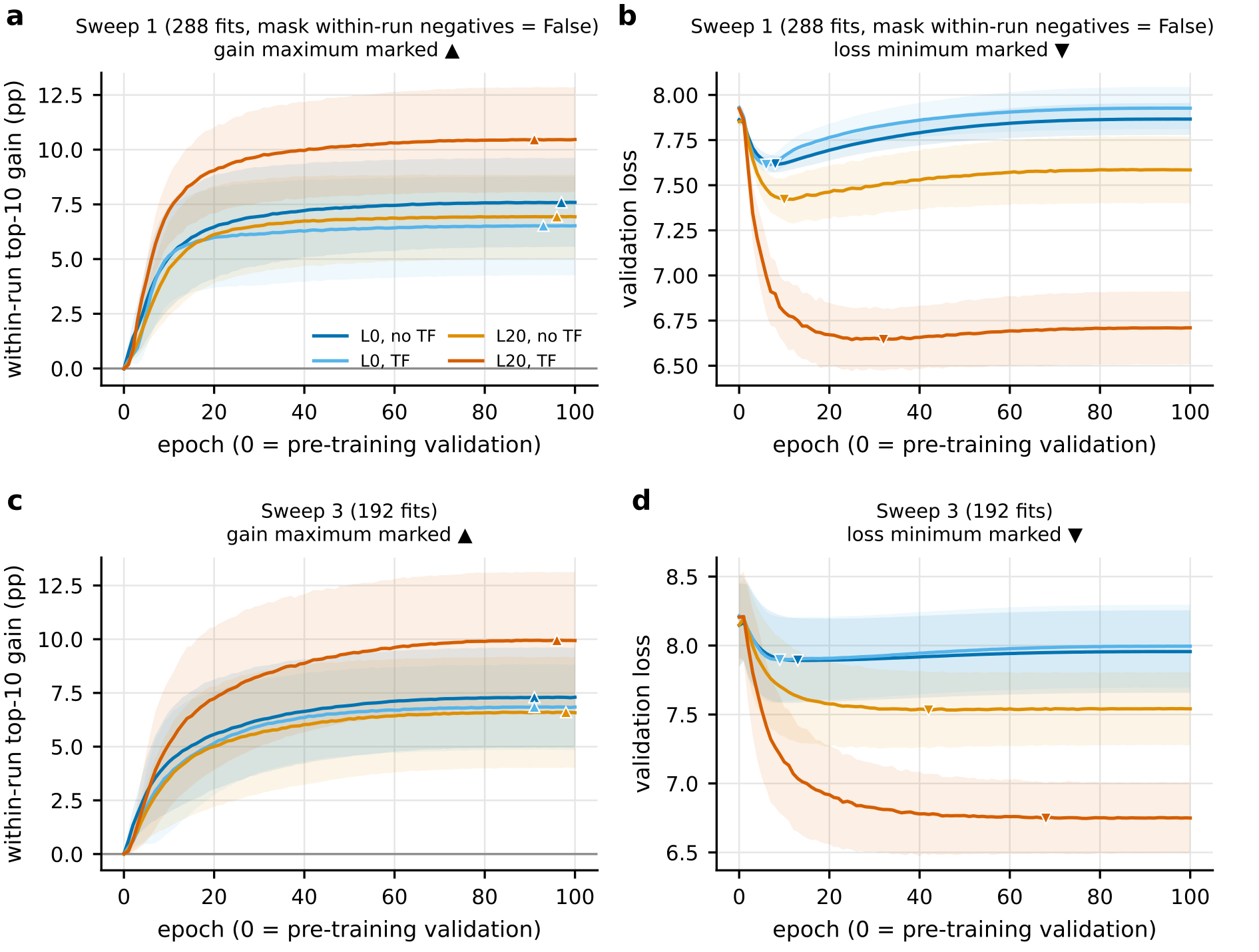}
  \caption{\textbf{Training dynamics: The validation-loss minimum and the within-run-gain maximum do not coincide, which is why checkpoints are selected on the gain.} \textbf{(a, b)} Sweep 1; \textbf{(c, d)} Sweep 3. Left column: within-run retrieval gain against epoch. Right column: validation loss against epoch. Lines are means across fits within each of the four target-type $\times$ architecture configurations, bands are $\pm1$ SD, and the extremum of each group-mean curve is marked ($\blacktriangle$ gain maximum, $\blacktriangledown$ loss minimum). Epoch 0 is the pre-training validation pass, so every gain curve starts at ${\sim}0$ by construction. The Sweep 1 row is restricted to fits with within-run negative masking off: masking changes the number of terms in the contrastive softmax, so validation loss is on a different scale in the two masking conditions and averaging across them would be meaningless. The gain panel is restricted identically so both panels in the row describe the same set of fits. Sweep 3 holds masking off throughout, so no filter applies there. In the non-contextual configurations the validation loss reaches a minimum around epoch 7--13 and rises thereafter while the within-run gain keeps improving. Selecting checkpoints on validation loss would therefore pick early, weak checkpoints.}
  \label{fig:s1-training-dynamics}
\end{figure}

\begin{figure}[tp]
  \centering
  \includegraphics[alt={Histogram of within-run gains for all fits against the pre-training null, and box plots split by configuration},width=\textwidth]{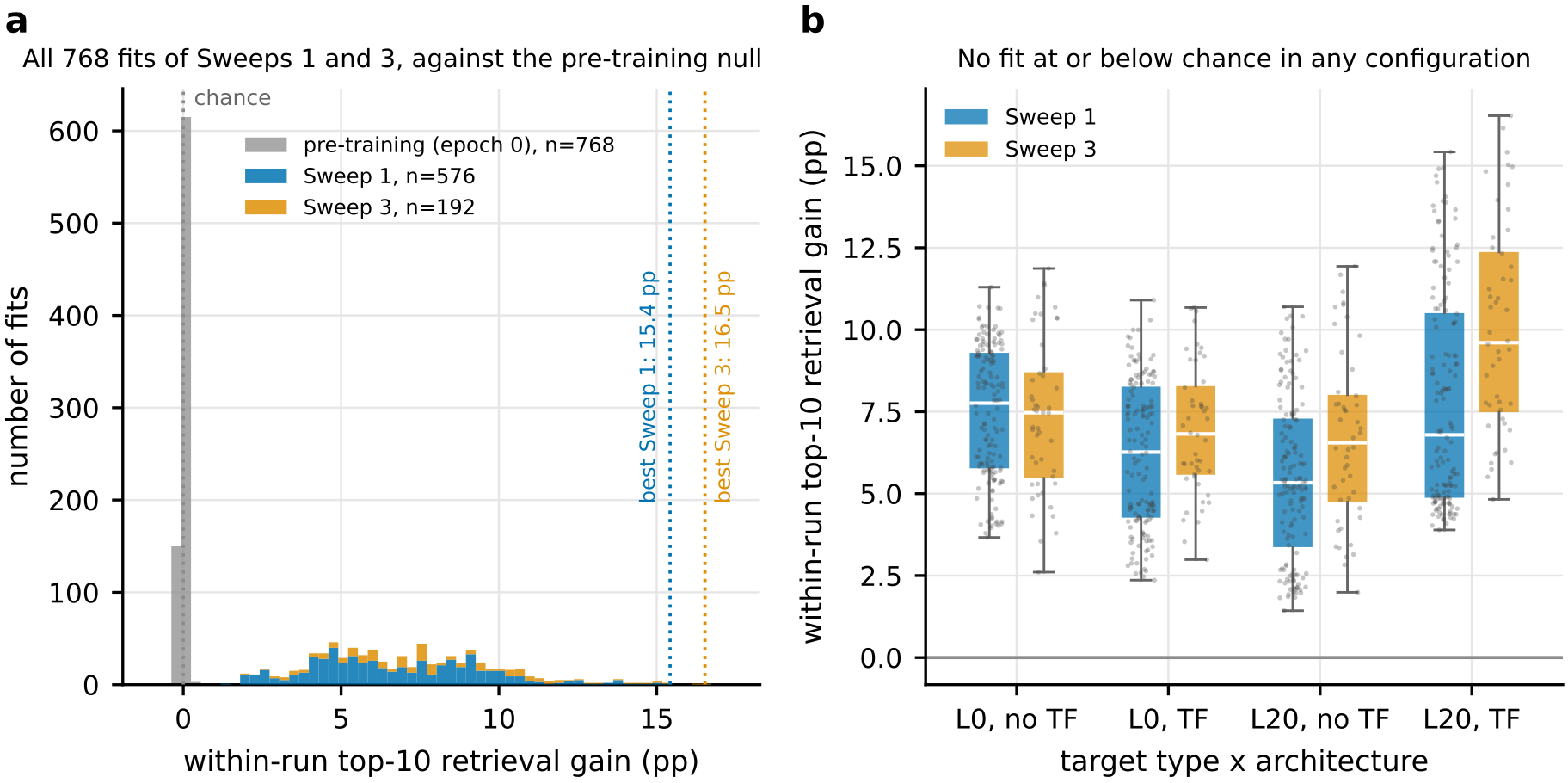}
  \caption{\textbf{Distribution of the primary metric across all fits, against its pre-training null.} \textbf{(a)} Within-run retrieval gain at the selected checkpoint for all 768 fits of Sweeps 1 and 3, stacked by sweep (colour), overlaid on the pre-training (epoch 0) values of the same fits (grey). The grey distribution is the empirical null for this metric: the model has been initialised but not trained, so any departure from zero is the metric's own noise given five evaluation pool draws and ten permutations per pool. The two distributions do not overlap: the pre-training values lie within $\pm0.2$\,pp of zero, while the selected-checkpoint values run from 1.4 to 16.5\,pp, with medians of 6.3\,pp (Sweep 1) and 7.5\,pp (Sweep 3). The best single fit of each sweep is marked (15.4\,pp in Sweep 1, 16.5\,pp in Sweep 3); both are maxima over fits as well as over epochs and should be read as upper bounds. \textbf{(b)} The same values split by target type $\times$ architecture as box plots (median, quartiles, whiskers at $1.5 \times$ IQR) with individual fits jittered behind. No fit in any of the four configurations falls at or below chance, so the decoding effect is present everywhere in the explored hyperparameter space and the sweeps determine its magnitude rather than its existence. Chance is zero for every value shown.}
  \label{fig:s2-gain-distribution}
\end{figure}

\section{Word presentations surviving the amplitude criterion}
\label{sec:s8}

The 100\,\muV{} amplitude criterion (Section~\ref{sec:preprocessing}) is applied to the analysis window plus the temporal-shift margin (for the data augmentation hyperparameter), so a longer window, a different lock point (onset, offset, centre) or a larger augmentation shift can result in a different number of trials being excluded. The inclusion count is therefore a property of the analysis cell.

\begin{table}[htb]
  \centering
  \caption{Word presentations surviving the 100\,\muV{} amplitude criterion, by analysis-window parameter (mean across the fits in each cell; range in parentheses where it varies).}
  \label{tab:s8-inclusion}
  \footnotesize
  \setlength{\tabcolsep}{4pt}
  \begin{tabular}{llcccccc}
    \toprule
    Factor & Level & Fits & Presented & Included & Excluded & \makecell[c]{Inclusion\\rate} & Sweep \\
    \midrule
    Window lock & onset & 64 & 240{,}141 & 220{,}787 (218{,}530--223{,}044) & 19{,}354 & 91.0--92.9\% & Sweep 1 \\
    & centre & 64 & 240{,}141 & 221{,}334 (219{,}018--223{,}651) & 18{,}806 & 91.2--93.1\% & Sweep 1 \\
    & offset & 64 & 240{,}141 & 221{,}832 (219{,}670--223{,}995) & 18{,}308 & 91.5--93.3\% & Sweep 1 \\
    Temporal shift & 0 & 96 & 240{,}141 & 223{,}563 (223{,}044--223{,}995) & 16{,}578 & 92.9--93.3\% & Sweep 1 \\
    & $\pm100$\,ms & 96 & 240{,}141 & 219{,}073 (218{,}530--219{,}670) & 21{,}068 & 91.0--91.5\% & Sweep 1 \\
    Window length & 0.5\,s & 64 & 240{,}141 & 222{,}157 (220{,}498--223{,}790) & 17{,}984 & 91.8--93.2\% & Sweep 3 \\
    & 0.7\,s & 64 & 240{,}141 & 221{,}056 (219{,}651--222{,}443) & 19{,}085 & 91.5--92.6\% & Sweep 3 \\
    & 0.9\,s & 64 & 240{,}141 & 216{,}122 (213{,}808--218{,}429) & 24{,}019 & 89.0--91.0\% & Sweep 3 \\
    Temporal shift & 0 & 96 & 240{,}141 & 221{,}134 (217{,}487--223{,}790) & 19{,}007 & 90.6--93.2\% & Sweep 3 \\
    & $\pm50$\,ms & 96 & 240{,}141 & 218{,}422 (213{,}808--221{,}295) & 21{,}718 & 89.0--92.2\% & Sweep 3 \\
    Channel set & all 19 & 96 & 240{,}141 & 219{,}350 (213{,}808--223{,}044) & 20{,}791 & 89.0--92.9\% & Sweep 3 \\
    & O1/O2/T5/T6 removed & 96 & 240{,}141 & 220{,}207 (214{,}764--223{,}790) & 19{,}934 & 89.4--93.2\% & Sweep 3 \\
    \bottomrule
  \end{tabular}
\end{table}

\end{document}